\documentclass[lettersize,journal]{IEEEtran}
\usepackage{amsmath,amsfonts}
\usepackage{algorithmic}
\usepackage{algorithm}
\usepackage{array}
\usepackage[caption=false,font=normalsize,labelfont=sf,textfont=sf]{subfig}
\usepackage{textcomp}
\usepackage{stfloats}
\usepackage{url}
\usepackage{verbatim}
\usepackage{graphicx}
\usepackage{cite}
\usepackage{subcaption}
\usepackage{bm}
\usepackage{cancel}
\usepackage{nicefrac}  
\usepackage{xcolor}
\usepackage{microtype}
\usepackage{multirow}
\usepackage{booktabs}
\usepackage{makecell}

\newcommand{\xt}[1]{{\textcolor[rgb]{0,0,0.0}{#1}}}

\begin{document}

\title{Human Motion Prediction via \\ Continual Prior Compensation}

\author{Jianwei Tang, Jian-Fang Hu, Tianming Liang, Xiaotong Lin, Jiangxin Sun, Wei-Shi Zheng, Jianhuang Lai
\thanks{J. Tang, J.-F. Hu, T. Liang, X. Lin, J. Sun, W.-S. Zheng, and J. Lai are with the School of Computer Science, Sun Yat-sen University, Guangzhou, China. J.-F. Hu, W.-S. Zheng, and J. Lai are also with the Guangdong Province Key Laboratory of Information Security Technology and the Key Laboratory of Machine Intelligence and Advanced Computing (Sun Yat-sen University), Ministry of Education, China. The corresponding author is J.-F. Hu. E-mail: \{tangjw7, liangtm, linxt29, sunjx5\}@mail2.sysu.edu.cn, wszheng@ieee.org, \{hujf5, stsljh\}@mail.sysu.edu.cn.}}

 \markboth{IEEE Transaction on Pattern Analysis and Machine Intelligence, 2025} 
{Tang \MakeLowercase{\textit{et al.}}: Human Motion Prediction via Continual Prior Compensation}


\maketitle

\begin{abstract}
Human Motion Prediction (HMP) aims to predict future human poses at different moments according to observed past motion sequences. Previous approaches mainly treated the prediction of different temporal moments as a single prediction task and learned the predictions of varied moments simultaneously, which would encounter a main limitation: the learning of short-term predictions (referring to ``near-future'' prediction) could be hindered by the predictions of long-term (referring to ``far-future'' prediction) motions. In this paper, we develop a novel temporal continual learning framework called Continual Prior Compensation (CPC) to progressively train HMP models, in which we divide the prediction task of motions corresponding to varied temporal moments into several subtasks and train the model in a multi-stage manner. To mitigate the prior information forgetting in the progressive training, we further introduce a learnable random variable Prior Compensation Factor (PCF) to explicitly measure the prior knowledge loss. We theoretically show that the PCF can be efficiently learned together with the model parameters by minimizing a reasonable upper bound of the objective function. The proposed CPC is further enhanced to estimate the prior information loss for each subtask and a new framework called Continual Prior Compensation++ (CPC++) with Fine-Grained Prior Compensation Factor (FGPCF) is finally developed. Our CPC and CPC++ frameworks are quite flexible and can be easily integrated with different HMP backbone models and adapted to various datasets and applications. Extensive experiments on three HMP benchmark datasets using multiple SOTA HMP backbones (PGBIG, siMLPe, MotionMixer, and LTD) demonstrate the effectiveness and flexibility of our frameworks.
\end{abstract}


\section{Introduction}

\IEEEPARstart{H}{uman} Motion Prediction (HMP) aims to predict future poses at varied temporal moments based on the observed motion sequences. The accurate prediction of human motion plays a vital role in many applications, such as autonomous driving, human-robot interaction, and security monitoring, enabling the anticipation and mitigation of risks. This task is challenging due to its requirement for predicting multiple moments, including short-term predictions for the ``near-future'' and long-term predictions for the ``far-future''. 

Previous approaches addressed this task by autoregressively forecasting using recurrent neural networks (RNNs) and transformer architectures \cite{aksan2021spatio,chiu2019action,corona2020context,ghosh2017learning,gopalakrishnan2019neural,gui2018adversarial,gui2018few,guo2019human,liu2019towards,sang2020human, sun2021action}, or parallelly generating all frames with graph convolution networks (GCNs) \cite{aksan2019structured,cui2021towards,cui2020learning,li2020dynamic,liu2020multi,mao2020history,mao2019learning,ma2022progressively}. These methods employed the one-stage training strategy to acquire the model to learn both short and long-term prediction simultaneously. However, the long-term motion prediction is more challenging since the future motion can vary greatly (i.e., the prediction space is large), which would increase the uncertainty and ambiguity of future prediction. If we train short-term and long-term predictions together, the fitting of high-uncertainty long-term prediction will dominate the learning process, which hinders the learning of short-term predictions.

\begin{figure*}[ht]
  \centering
  \includegraphics[width=0.99\textwidth]{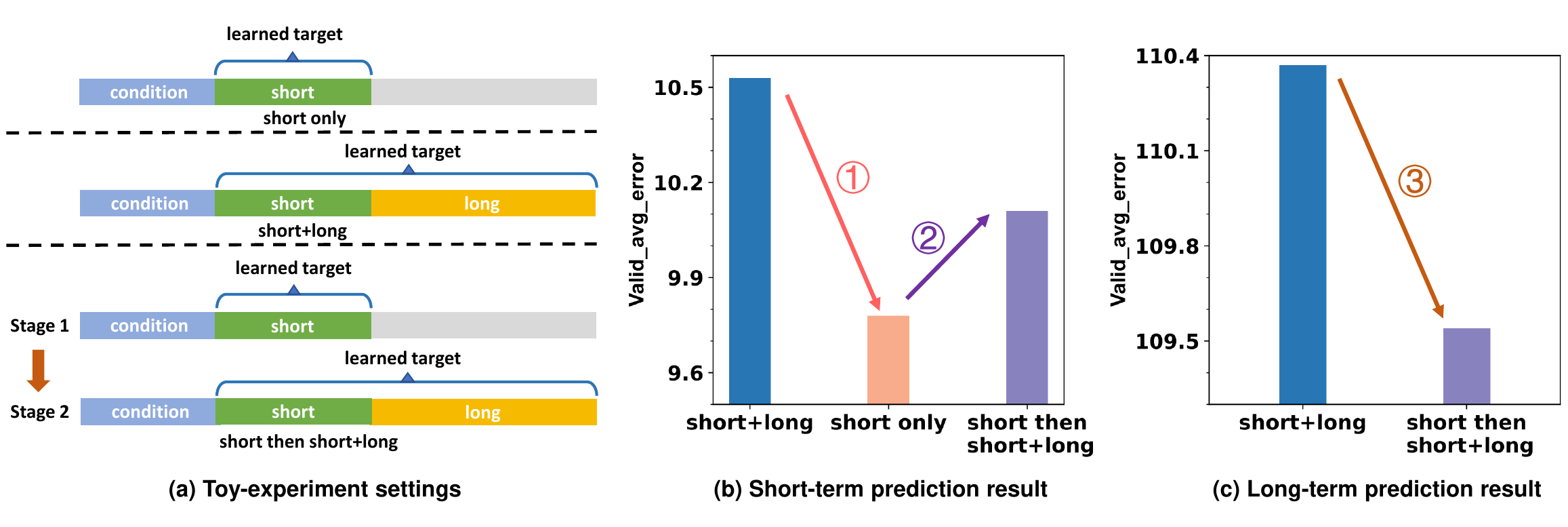}
  \caption{Preliminary experiment results of three different prediction settings (lower values indicate better performance). (a) represents the experimental settings. (b) shows the short-term prediction results (predicting the 2-nd frame), while (c) illustrates the long-term prediction results (predicting the 25-th frame).}
  \label{fig.1.1.1}
\end{figure*}

This motivates us to exploit proper training strategies to better learn the prediction for both short and long-term prediction. To this end, we conduct preliminary experiments with three settings: \textit{short+long}, \textit{short only} and \textit{short then short+long}, as illustrated in Figure 1(a). We first observe that ``short only'' outperforms ``short+long'' on short-term prediction (Figure \ref{fig.1.1.1}(b)), which implies that the joint learning of all timestamps is harmful to short-term prediction due to the fitting of high-uncertainty long-term prediction. This is intuitive and thus we can use a progressive learning approach, where the model is trained to predict increasing numbers of frames over multiple training stages, e.g., starting with 5 frames in the first stage and 10 frames in the second stage, and so on. However, we also observe that ``short then short+long'' performs worse than ``short only'' by a considerable margin for short-term prediction (Figure \ref{fig.1.1.1}(b)), which demonstrates the joint learning of short-term and long-term prediction in subsequent stages results in knowledge forgetting for short-term prediction. At the same time, we observe that ``short then short+long" outperforms ``short+long'' on long-term prediction in Figure \ref{fig.1.1.1}(c), which implies the knowledge learned in short-term prediction can serve as prior facilitating the learning of ``far-future'' prediction.

Inspired by the above analysis, we propose the Continual Prior Compensation (CPC) framework, which is a multi-stage training framework that alleviates constraints posed by long-term prediction on short-term prediction and effectively utilizes prior information from short-term prediction. Specifically, we divide the future unobserved sequence into several segments and regard the prediction of each segment as a subtask. We define the training process as multiple stages and progressively increase the number of prediction segments across the stages. This allows the model to leverage the prior knowledge acquired from earlier stages for predicting the subsequent ones. Upon completion of each stage's training, the learned prior knowledge is saved in the model parameters. However, in this process, the learning in subsequent stages could gradually eliminate the prior knowledge learned from former stages. To overcome the prior forgetting problem, we further introduce the Prior Compensation Factor (PCF), which is a learnable random variable that encodes the prior forgetting for all of the previous subtasks. However, directly learning the PCF and model parameters together is challenging. To this end, we further derive a reasonable upper bound of the objective function, which can be efficiently optimized.

In this work, we further introduce the Fine-Grained Prior Compensation Factor (FGPCF) and propose the Continual Prior Compensation++ (CPC++) framework to address the prior information forgetting problem more effectively. The FGPCF is defined to measure the prior information loss for each individual subtask. Similar to CPC, the enhanced CPC++ is also a multi-stage training approach, which intends to optimize the FGPCF and model parameters simultaneously. Our theoretical derivation shows that the CPC++ can be efficiently optimized by minimizing a reasonable upper bound function. 

We evaluate our method by conducting experiments on three popular HMP benchmarks (Human3.6M \cite{ionescu2013human3}, CMU-MoCap and 3DPW Dataset\cite{von2018recovering}) by integrating CPC and CPC++ with several HMP backbones (LTD \cite{mao2019learning}, MotionMixer(MM) \cite{bouazizi2022motionmixer}, siMLPe \cite{guo2023back} and PGBIG \cite{ma2022progressively}). Our results demonstrate that the proposed CPC and CPC++ training frameworks are flexible and can be easily integrated with various HMP backbones or adapted to different datasets. Additionally, the experimental results also indicate that the proposed PCF and FGPCF effectively mitigate the loss of prior knowledge, leading to improved performance in the HMP backbones.

In summary, our main contributions are threefold: 1) We identify certain limitations in existing HMP models and propose two novel multi-stage training strategies called Continual Prior Compensation and Continual Prior Compensation++ to enhance the training of the human motion prediction model. \xt{To the best of our knowledge, we are the first to develop such multi-stage training framework to progressively train human motion prediction models in a temporal continual learning manner.} 2) We introduce the Prior Compensation Factor and Fine-Grained Prior Compensation Factor to tackle the forgetting problem of prior knowledge, which can be learned jointly with the prediction model parameters. 3) We theoretically derive an easily optimized and reasonable objective function for effective optimization. We also present an extensive experimental analysis of four backbone HMP models and three benchmark datasets to illustrate the effectiveness of the proposed approach.

\xt{A preliminary version of the current work was reported in [42], which uses a learnable random variable (PCF) to encode the prior forgetting for all of the previous subtasks. In this work, we have significantly extended our framework in the following three aspects. First, we introduce the Fine-Grained Prior Compensation Factor (FGPCF) containing multiple learnable random variables, each of which measures the prior information loss for the corresponding prediction subtask. Second, we propose the Continual Prior Compensation++ (CPC++) framework, which intends to optimize the FGPCF and model parameters simultaneously. And an efficient optimization algorithm is obtained through theoretical derivation. Third, we conduct a more comprehensive comparative analysis of our framework to demonstrate the advantages of utilizing the FGPCF for prior loss estimation and report improved results on three datasets.}
\section{Related Work}
Our work is closely related to one-stage training and multi-stage training approaches for human motion prediction, and continual learning, which have been extensively investigated in the community. In the following, we will provide a brief review of these works.

\noindent
\textbf{One-stage training approaches for HMP.} 
Motivated by natural language processing, many researchers have adopted autoregressive models to process temporal sequences of human poses, which include RNNs, Long Short-Term Memory Networks (LSTMs) \cite{hochreiter1997long} and Transformer \cite{vaswani2017attention}. For instance, ERD \cite{fragkiadaki2015recurrent} combined LSTMs with an encoder-decoder to model the temporal aspect, while Jain \textit{et al.} \cite{jain2016structural} proposed Structural-RNN to capture spatiotemporal features of human motion. Martinez \textit{et al.} \cite{martinez2017human} applied a sequence-to-sequence architecture for modeling the human motion structure. Aksan \textit{et al.} \cite{aksan2021spatio} used Transformer to predict future poses autoregressively. Sun \textit{et al.} \cite{sun2021action} designed a query-read process to retrieve motion dynamics from the memory bank. Lucas \textit{et al.} \cite{lucas2022posegpt} proposed a GPT-like\cite{radford2018improving} autoregressive method to generate human poses. However, autoregressive methods are difficult to train and suffer from error accumulation.

Some researchers employed parallel prediction methods to address HMP problem \cite{aksan2019structured,cui2021towards,cui2020learning,li2020dynamic,liu2020multi,mao2020history,mao2019learning,ma2022progressively}. Many works (\cite{lebailly2020motion,li2021symbiotic,li2020dynamic}) used GCN to encode feature or to decode it, which associates different joints' information. Mao \textit{et al.} \cite{mao2019learning} viewed a pose as a fully connected graph and used GCN to extract hidden information between any pair of joints. Martinez \textit{et al.} \cite{martinez2021pose} devised a transformer-based network to predict human poses. Sofianos \textit{et al.} \cite{sofianos2021space} proposed a method to extract spatiotemporal features using GCNs. And Ma \textit{et al.} \cite{ma2022progressively} tried to achieve better prediction results by progressively generating better initial guesses. Xu \textit{et al.} \cite{xu2022diverse} used multi-level spatial-temporal anchors to achieve diverse predictions. \xt{Wan et al. \cite{wan2025ggmotion} proposed GGMotion to model human topology in groups which better leverages dynamics and kinematics priors. Ding et al. \cite{ding2025ksof} proposed to capture temporal and spatial dependencies via a kinematic temporal convolutional network and spatial graph convolutional networks, respectively. Li et al. \cite{li2022skeleton} constructed adaptive graph scattering across various body parts to capture motion dynamics. These work intended to facilitate motion prediction by developing stronger motion representations, which differs significantly from our objective.} 

\noindent
\textbf{Multi-stage training approaches for HMP.} Some researchers proposed multi-stage prediction methods to handle this task. Yuan \textit{et al.} \cite{yuan2020dlow} utilized a two-stage conditional Variational Autoencoder (cVAE) \cite{kingma2013auto} model for diverse human motion prediction. Barquero \textit{et al.} \cite{barquero2023belfusion} introduced the BeLFusion model, a two-stage latent diffusion model for the diverse HMP task. They utilized a two-stage learning process, where the first stage learns the VAE generation process, and the second stage focuses on enhancing the diversity of sampling. However, it is worth noting that these models were all acquired to learn both short-term and long-term prediction simultaneously during the training process. It is significantly different from our multi-stage training framework which progressively trains the short-term and long-term predictions in different stages.

\noindent
\textbf{Continual learning.}
Although Deep Neural Networks (DNNs) have demonstrated impressive performance on specific tasks, their limitations in handling diverse tasks hinder their broader applicability. The most severe problem is the 
catastrophic forgetting problem when simply applying the DNNs to multiple tasks. Therefore, some researchers introduced the concept of Continual Learning (CL)\cite{ring1997child} to DNNs to ensure that models retain the knowledge of previous tasks while learning new tasks. Kirkpatrick \textit{et al.} \cite{kirkpatrick2017overcoming} proposed the Elastic Weight Consolidation (EWC) method to overcome the catastrophic forgetting problem and improve the performance of multi-task problems. Shin \textit{et al.} \cite{shin2017continual} introduced a method that addresses catastrophic forgetting in sequential learning scenarios by using a generative model to replay data from past tasks during the training of new tasks. It is important to note that the traditional CL approaches do not account for temporal correlation and cannot leverage data from previous tasks.

\section{Continual Prior Compensation}
In this part, we introduce the details of the proposed Continual Prior Compensation (CPC) framework. We first provide a problem formulation in Section \ref{sec:3.1}. Section \ref{sec:3.2} introduces the formal modeling and derivation. Finally, the optimization strategy is presented in Section \ref{sec:3.3}.

\subsection{Problem Formulation}
\label{sec:3.1}
Human motion prediction aims to predict future motion sequences conditioned on the previously observed motion sequences. Formally, let us use $\mathbf{X}_{1:T_h}=[\mathbf{X}_1, \mathbf{X}_2,\cdots, \mathbf{X}_{T_h}]\in \mathbb{R}^{J\times D\times T_h} $ to denote the observed motion sequence of length $T_h$ where $\mathbf{X}_i$ indicates motion at temporal location $i$. Note that $J$ is the number of joints for each pose, and $D$ is the dimension of coordinates. $\mathbf{X}_{T_{h}+1:T_{h}+T_p}=[\mathbf{X}_{T_{h}+1}, \mathbf{X}_{T_{h}+2},\cdots, \mathbf{X}_{T_{h}+{T_p}}]\in \mathbb{R}^{J\times D\times T_p}$ represents the motion sequence to be predicted. $T_p$ is the length of the predicted sequence. It can be regarded as a composition of multiple sequential prediction subtasks, which involves predicting motions at varied future moments. 

With the above denotations, the multiple timestamp prediction problem (i.e., HMP prediction) can be formulated as solving the following optimization problem:
\begin{equation}
\label{eq:optimizationProblem}
  \begin{split}
    \bm{\theta}^* =& \mathop{\arg\max}_{\bm{\theta}} P(\mathbf{X}_{T_{h}+1:T_{h}+T_p}|\mathbf{X}_{1:T_h};\bm{\theta}), \\
  \end{split}
\end{equation}
which means that our target is to find the optimal model that maximizes Equation (\ref{eq:optimizationProblem}). Here, $\bm{\theta}$ is the model parameters to be learned.

Previous approaches mainly formulated models to learn both short and long-term prediction simultaneously in a one-stage manner during the training process. However, one-stage training methods would encounter a main limitation: the learning of short-term predictions (referring to “near-future” prediction) could be hindered by the predictions of long-term (referring to “far-future” prediction) motions. In this work, we propose to conduct the model training in a multi-stage learning manner, in which the model is progressively trained by fitting ongoing prediction motion sequences. To this end, we partition the entire prediction interval into several consecutive segments and progressively tune the model to additionally fit new motion segments, which have been pre-trained with previous segments. 

Specifically, we initially split the future sequence into $K$ segments with temporal boundaries $T_{1}, T_{2}, \cdots, T_{K}$, where $T_{K}=T_h+T_p$. And we denote the prediction of segment $k$ as task $Z_k$, which can be expressed as follows:
\begin{itemize}
  \item Task $Z_{1}:\quad \mathbf{X}_{1:T_h} \rightarrow \mathbf{X}_{T_{h}+1:T_{1}}$
  \item Task $Z_{2}:\quad \mathbf{X}_{1:T_h} \rightarrow \mathbf{X}_{T_{1}+1:T_{2}}$
  \\ $\cdots$
  \item Task $Z_{K}:\quad \mathbf{X}_{1:T_h} \rightarrow \mathbf{X}_{T_{K-1}+1:T_{K}}$
\end{itemize}
To be specific, the target of task $Z_1$ is to predict $\mathbf{X}_{T_{h}+1:T_{1}}$ conditioned on $\mathbf{X}_{1:T_h}$, and task $Z_k$ aims to predict $\mathbf{X}_{T_{k-1}+1:T_{k}}$ with $\mathbf{X}_{1:T_h}$ as condition. Therefore, based on the Bayesian formulation, Equation (\ref{eq:optimizationProblem}) can be expressed as:
\begin{equation}
\label{eq:3.2.1}
  \begin{split}
    P(Z_{1}Z_{2}\cdots &Z_{K};\bm{\theta})
                  = P(Z_{K}|Z_{1}Z_{2}\cdots Z_{K-1};\bm{\theta})\\
                  &P(Z_{K-1}|Z_{1}Z_{2}\cdots Z_{K-2};\bm{\theta})\cdots P(Z_{1};\bm{\theta}). \\
  \end{split}
\end{equation}
Our target is to maximize the probability defined in Equation (\ref{eq:3.2.1}) which means finding an optimal model to accomplish all tasks. In the following, we denote ``$Z_{1}Z_{2}\cdots Z_{k}$'' as ``$Z_{1:k}$'' for simplicity.

In this way, we can train the model in a multi-stage manner, which facilitates the utilization of prior information from the previous tasks for predicting the subsequent tasks. Since the model is trained to optimize the prediction tasks $Z_{1:k-1}$ in training stage $S_{k-1}$, the knowledge of tasks $Z_{1:k-1}$ can be implicitly involved in its well-trained parameters $\bm{\theta}^*_{k-1}$. Initializing $\bm{\theta}_{k}$ as $\bm{\theta}^*_{k-1}$ can exploit the prior knowledge learned in previous tasks to assist the prediction of the next task. However, the change of the optimization objective in different training stages could also bring about the knowledge-forgetting problem due to the changing of parameters $\bm{\theta}_{k}$.

\subsection{Objective of Continual Prior Compensation}
\label{sec:3.2}
Here, we introduce the Prior Compensation Factor (PCF), which explicitly measures the knowledge forgotten in the multi-stage training.

\subsubsection{\textbf{Prior Compensation Factor}}
\label{sec:4.2}
We define the Prior Compensation Factor (PCF) as a random variable $\alpha_{Z_{1:k-1} \rightarrow Z_k}$, which estimates the extent of forgotten knowledge when utilizing prior knowledge from tasks $Z_{1:k-1}$ to predict task $Z_k$. Specifically, the PCF can be formulated as:
\begin{equation}
\label{eq:3.2.2}
\begin{split}
    \alpha_{Z_{1:k-1}\rightarrow Z_{k}} &= P(Z_{k}|Z_{1:k-1};\bm{\theta}) - P(Z_{k}|\hat{Z}_{1:k-1};\bm{\theta}),\\
    &\quad k\in\{2,3,\cdots,K\}.
\end{split}
\end{equation}
Here, $\hat{Z}_{1:k-1}$ is regarded as the prior knowledge that is reserved and can be still provided for predicting task $Z_k$. So $\hat{Z}_{1:k-1}$ initially represents the prior knowledge reserved in $\bm{\theta}^*_{k-1}$ in every stage $S_k$ and would get corrupted gradually during training. Therefore, $P(Z_{k}|\hat{Z}_{1:k-1};\bm{\theta})$ indicates the learning prediction ability of a new task using corrupted prior knowledge. While $P(Z_{k}|Z_{1:k-1};\bm{\theta})$ represents the most ideal case, where the current prediction task $Z_{k}$ can fully leverage the prior information provided by previous prediction tasks $Z_{1:k-1}$. Consequently, the loss of the prior knowledge is non-negative, implying that $0\le \alpha_{Z_{1:k-1}\rightarrow Z_{k}} \le P(Z_{k}|Z_{1:k-1};\bm{\theta}) - P(Z_{k}|\hat{Z}_{1:k-1};\bm{\theta})\le 1 - P(Z_{k}|\hat{Z}_{1:k-1};\bm{\theta})$. The larger the value of $\alpha_{Z_{1:k-1}\rightarrow Z_{k}}$ is, the more prior knowledge is forgotten. Specifically, we can observe that $\alpha_{Z_{1:k-1}\rightarrow Z_{k}}=0$ when $ P(Z_k|Z_{1:k-1};\bm{\theta})=P(Z_k|\hat{Z}_{1:k-1};\bm{\theta}) $, which implies that all the prior knowledge of previous tasks is completely exploited although the $\bm{\theta}$ changes. It is worth noting that the value of $\alpha_{Z_{1:k-1}\rightarrow Z_{k}}$ could be varied for different training samples.

\subsubsection{\textbf{Optimization Objective}}
\label{sec:4.3}
By substituting Equation (\ref{eq:3.2.2}) into Equation (\ref{eq:3.2.1}) and taking the negative logarithm, we can obtain:
\begin{equation}
\label{eq:3.2.4}
\begin{split}
  -\ln &P(Z_{1:k};\bm{\theta})= -\ln P(Z_{1};\bm{\theta})  \\
  &-\sum_{i=2}^{k}\ln(P(Z_{i}|\hat{Z}_{1:i-1};\bm{\theta})+\alpha_{Z_{1:i-1}\rightarrow Z_{i}}) .
\end{split}
\end{equation}
Our target turns to minimize $-\ln P(Z_{1:k};\bm{\theta})$ with respect to the model parameter $\bm{\theta}$ and prior compensation factors $\{\alpha_{Z_{1:i-1}\rightarrow Z_{i}}, i=2,3,\cdots, k\}$. Optimizing Equation (\ref{eq:3.2.4}) directly is challenging due to the difficulty of determining the value of PCF during training. However, by applying Lemma 3.1 (details provided in the following), we can obtain an upper bound for $-\ln P(Z_{1:k};\bm{\theta})$, which can be expressed as:
\begin{equation}
  \label{eq:3.2.5}
  \begin{split}
    UB =& \sum_{i=2}^{k}((1-\alpha_{Z_{1:i-1}\rightarrow Z_{i}})(-\ln P(Z_{i}|\hat{Z}_{1:i-1};\bm{\theta}))  \\
                        & + (1-\alpha_{Z_{1:i-1}\rightarrow Z_{i}})\ln(1-\alpha_{Z_{1:i-1}\rightarrow Z_{i}}) \\
                        &+ \ln(1+\alpha_{Z_{1:i-1}\rightarrow Z_{i}})) -\ln P(Z_{1};\bm{\theta}) .\\
  \end{split}
\end{equation}
Hence, we can turn to minimize the upper bound indicated by Equation (\ref{eq:3.2.5}). It is worth noting that in the optimization objective, $\alpha_{Z_{1:i-1}\rightarrow Z_{i}}$ serves as a factor to control the weights of different tasks. When the prior knowledge loss of previous tasks becomes severe, the weight of the current task's loss will be smaller, which mitigates the prior knowledge loss of previous tasks and thus compensates for the lost prior knowledge. We would like to point out that the largest difference between $-\ln P(Z_{1:k};\bm{\theta})$ and the upper bound $UB$ would not exceed $\ln(\nicefrac{3}{2})*(k-1)$ in the case of $P(Z_{i}|\hat{Z}_{1:i-1};\bm{\theta}) \ge \nicefrac{1}{2}, i\in \{2,3,\cdots, k\}$, which would be demonstrated in Lemma 3.2.

\textit{\textbf{Lemma 3.1.}} For $0\leq a \leq 1-b$ and $0 < b \leq 1$, the inequality $-\ln(a + b) \leq (1-a)(-\ln b) + (1-a)\ln(1-a) + \ln(1+a)$ holds. The equality holds if and only if $a=0$.

\textit{Proof:} Let's consider the following function:
\begin{equation}
\label{eq:Ga}
\begin{split}
    G(a) &= \ln(a+b) - (1-a)\ln b \\
         &+ (1-a)\ln(1-a) + \ln(1+a).
\end{split}
\end{equation}
By defining $G_1(a)=\ln(a+b) - (1-a)\ln b$ and $G_2(a)=(1-a)\ln(1-a) + \ln(1+a)$, we can observe that $G(a)=G_1(a)+G_2(a)$.

Regarding $G_1(a)$, its derivative and second derivative can be computed as follows:
\begin{equation}
\label{eq:Ga1_d12}
\begin{split}
    \dot{G_1}(a) &= \frac{1}{a+b} + \ln b \\
    \ddot{G_1}(a) &= -\frac{1}{(a+b)^2}.
\end{split}
\end{equation}
For the case of $b \ge \nicefrac{1}{e}$, since $\ddot{G_1}(a) < 0$ and $0 \leq a \leq 1-b$, we can obtain that $\dot{G_1}(a) \ge \dot{G_1}(1-b) = 1+\ln b \ge 0$. Therefore, $G_1(a)\ge G_1(0)=0$. For the case of $0 < b < \nicefrac{1}{e}$, we can easily obtain that $\dot{G_1}(1-b) = 1+\ln b < 0$ and $\dot{G_1}(0) = \nicefrac{1}{b} + \ln b > 0$. Hence, by considering the monotonicity of $\dot{G_1}(a)$, we can conclude that there exists an $ a_0\in (0,1-b)$ such that $\dot{G_1}(a) \ge 0$ for $a\in [0,a_0]$ and $\dot{G_1}(a) < 0$ for $a\in (a_0, 1-b]$. As a result, we can get $G_1(a) \ge \min (G_1(0), G_1(1-b))$. Since $G_1(0) = 0$ and $G_1(1-b) = -b\ln b \ge 0$, we can finally have $G_1(a) \ge 0$, the equality holds if only if $a = 0$.

In the following, we show that $G_2(a)\geq 0$ holds. The first derivative $\dot{G_2}(a)$ and second derivative $\ddot{G_2}(a)$ are given by:
\begin{equation}
    \label{eq:Ga2_d12}
    \begin{split}
        \dot{G_2}(a) &=  -\ln(1-a) -1 + \frac{1}{1+a}\\
        \ddot{G_2}(a) &= \frac{1}{1- a}  - \frac{1}{(1 + a)^2}.
    \end{split}
\end{equation}
Since $\ddot{G_2}(a) \ge 0$ for any $a \in[0,1-b]$, we can get $\dot{G_2}(a) \geq \dot{G_2}(0)=0$ and thus $G_2(a) \ge G_2(0) = 0$. 

With $G_1(a) \ge 0$ and $G_2(a)\ge 0$, we can obtain $G(a)=G_1(a)+G_2(a)\ge 0$ and conclude that $-\ln(a + b) \leq (1-a)(-\ln b) + (1-a)\ln(1-a) + \ln(1+a)$. Moreover, we can observe that equality holds if and only if $a=0$.

\textit{\textbf{Lemma 3.2.}} The absolute difference between the target objective (Equation (\ref{eq:3.2.4})) and the upper bound (Equation (\ref{eq:3.2.5})) is not larger than $\ln(\nicefrac{3}{2})*(k-1)$ when $P(Z_{i}|\hat{Z}_{1:i-1};\bm{\theta}) \ge \nicefrac{1}{2}, i\in \{2,3,\cdots, k\}$. This bound is achieved when $P(Z_{k}|\hat{Z}_{1:k-1};\bm{\theta}) = \nicefrac{1}{2}$ and $\alpha_{Z_{1:i-1}\rightarrow Z_{i}} = \nicefrac{1}{2}, i\in \{2,3,\cdots, k\}$.

\textit{Proof:} For simplicity, let's denote $\alpha_i=\alpha_{Z_{1:i-1}\rightarrow Z_i}$ and $p_i=P(Z_i|\hat{Z}_{1:i-1};\bm{\theta})$, where $p_i\in[\nicefrac{1}{2},1], \alpha_i\in[0,1-p_i], i\in\{2,3,\cdots, k\}$. Then the difference between the upper bound (Equation (\ref{eq:3.2.5})) and the target objective (Equation (\ref{eq:3.2.4})) can be calculated as:
\begin{equation}
    \label{eq:DIF_ACUB}
    \begin{split}
        \tau = & UB - (-\ln P(Z_{1:k};\bm{\theta})) \\
        =& \sum_{i=2}^{k}(\ln(\alpha_i+p_i) - (1-\alpha_i)\ln p_i  \\
        &+ (1-\alpha_i)\ln(1-\alpha_i) + \ln(1+\alpha_i)) ) .\\
    \end{split}
\end{equation}
Note that each term in the summation of Equation (\ref{eq:DIF_ACUB}) has the same form, we denote it as $T(\alpha)$, where
\begin{equation}
\label{eq:3.11}
\begin{split}
    T(\alpha) &= \ln(\alpha+p) - (1-\alpha)\ln p \\
    &+ (1-\alpha)\ln(1-\alpha) + \ln(1+\alpha) .
\end{split}
\end{equation}
Then, we can turn to maximize $T(\alpha)$ in order to obtain the largest difference between the target objective and its upper bound.

The first derivative of $T(\alpha)$ can be calculated as:
\begin{equation}
\label{eq:3.12}
    \dot{T}(\alpha) = \frac{1}{\alpha+p} + \ln p -\ln(1-\alpha) -1 + \frac{1}{1+\alpha} .
\end{equation}
By defining $T_1(\alpha)=\nicefrac{1}{(\alpha+p)} + \ln p$ and $T_2(\alpha)= -\ln(1-\alpha) -1 + \nicefrac{1}{(1+\alpha)}$, we can observe that $\dot{T}(\alpha) = T_1(\alpha) + T_2(\alpha)$. The first derivative of $T_2(\alpha)$ is:
\begin{equation}
\label{eq:3.13}
  \dot{T_2}(\alpha) = \frac{1}{1- \alpha}  - \frac{1}{(1 + \alpha)^2} .
\end{equation}
Since $\dot{T_2}(\alpha) \ge 0$ for any $\alpha\in[0,1-p]$, we get $T_2(\alpha) \geq T_2(0)=0$. Furthermore, when $p\ge\nicefrac{1}{2}$, $ T_1(\alpha) > 0$. Therefore, $\dot{T}(\alpha) = T_1(\alpha) + T_2(\alpha) >0$, which means the maximum value of $T(\alpha)$ is:
\begin{equation}
  \label{eq:Large_gap}
  \begin{split}
    T(1-p) &= -p\ln p + p\ln p + \ln(2-p) \\
           &=  \ln(2-p) .
  \end{split}
\end{equation}
Since $p\ge\nicefrac{1}{2}$ and $\ln(2-p)$ decreases with $p$, $T(1-p)$ will not larger than $\ln(2-\nicefrac{1}{2})=\ln(\nicefrac{3}{2})$. Particularly, when $p=\nicefrac{1}{2}$ and $\alpha=1-p=\nicefrac{1}{2}$, this largest bound will be achieved. Hence, we can conclude that $T(\alpha) \le \ln(\nicefrac{3}{2})$, and the equality holds when $p=\alpha=\nicefrac{1}{2}$.

Thus, as $\tau = \sum_{i=2}^kT(\alpha_i)$ in Equation (\ref{eq:DIF_ACUB}), we obtain $\tau \le \sum_{i=2}^{k}\ln(\nicefrac{3}{2}) = \ln(\nicefrac{3}{2})*(k-1)$, which means the absolute difference between Equation (\ref{eq:3.2.5}) and Equation (\ref{eq:3.2.4}) will not be greater than $\ln(\nicefrac{3}{2})*(k-1)$. The equality holds if and only if $p_i= \alpha_i = \nicefrac{1}{2}, i\in \{2,3,\cdots, k\}$.

\subsection{Optimization Strategy}
\label{sec:3.3}
We train the model in a temporal continual learning (multi-stage) manner, in which a total of $K$ CPC stages are involved. In the first stage $S_1$, we aim to train the model for forecasting the motion in the foremost segment without estimating PCF. While for the stage $S_k, k\in \{2,3,\cdots, K\}$, we update our model for predicting motion segments of tasks $Z_{1:k}$ and measuring the PCF $\alpha_{Z_{1:k-1}\rightarrow Z_{k}}$. Once the model is completely trained, we turn to estimate the factors $\hat{\alpha}_{Z_{1:k-1}\rightarrow Z_{k}}$ for subsequent stages. This process is repeated until the final stage $S_K$ is finished. The algorithm flow is summarized in Algorithm \ref{alg3.1}. In the following, we elaborate on the learning process.

\noindent
\textbf{Learning of initial stage $S_1$.}
Following the implementations of previous methods \cite{kendall2018multi}, we can train the initial stage $S_{1}$ with mean suqared error (MSE) loss:
\begin{equation}
  \label{eq:3.2.13}
  \begin{split}
    \mathcal{L}_1 
                &= \sum_{i=T_{h}+1}^{T_{1}} {\left\lVert \mathbf{X}_{i} - \hat{\mathbf{X}}_{i} \right\rVert}^2,
  \end{split}
\end{equation}
where $\mathbf{X}_{i}$ and $\hat{\mathbf{X}}_{i}$ represent the ground truth and predicted motion of the $i$-th frames respectively. 

\noindent
\textbf{Learning of stage $S_k$.}
In stage $S_{k}$ ($k \ge 2$), we need to update the model parameters $\bm{\theta}$ corresponding to tasks $Z_{1:k}$ and the PCF $\alpha_{Z_{1:k-1}\rightarrow Z_{k}}$. In practice, PCF can be calculated by adding an MLP head to the backbone model. According to Equation (\ref{eq:3.2.5}), the loss function in this stage can be calculated as follows:
\begin{equation}
  \label{eq:3.2.15}
  \begin{split}
    \mathcal{L}_{k} &= (1-\alpha_{Z_{1:k-1}\rightarrow Z_{k}})\sum_{i=T_{k-1}+1}^{T_{k}} {\left\lVert \mathbf{X}_{i} - \hat{\mathbf{X}}_{i} \right\rVert}^2 \\
    &+ (1-\alpha_{Z_{1:k-1}\rightarrow Z_{k}})\ln(1-\alpha_{Z_{1:k-1}\rightarrow Z_{k}})\\
                &+ \ln(1+\alpha_{Z_{1:k-1}\rightarrow Z_{k}}) \\
                &+ \sum_{j=2}^{k-1}(1-\hat{\alpha}_{Z_{1:j-1}\rightarrow Z_{j}})\sum_{i=T_{j-1}+1}^{T_{j}} {\left\lVert \mathbf{X}_{i} - \hat{\mathbf{X}}_{i} \right\rVert}^2 + \mathcal{L}_{1}, \\
  \end{split}
\end{equation}
where the parameters $\hat{\alpha}_{Z_{1}\rightarrow Z_{2}},\cdots,\hat{\alpha}_{Z_{1:k-2}\rightarrow Z_{k-1}}$ are dertermined in the learning of previous stages. Once the model parameters for stage $S_{k}$ are determined, we then calculate $\hat{\alpha}_{Z_{1:k-1}\rightarrow Z_{k}}$ as: 
\begin{equation}
  \label{eq:4.1.3}
  \hat{\alpha}_{Z_{1:k-1}\rightarrow Z_{k}} = \frac{1}{M}\sum_{m=1}^{M}\hat{\alpha}^{m}_{Z_{1:k-1}\rightarrow Z_{k}},
\end{equation}
where $M$ represents number of samples and $\hat{\alpha}^{m}_{Z_{1:k-1}\rightarrow Z_{k}}$ is PCF estimated for the $m$-th sample.

We continue the CPC training process by progressively predicting tasks $Z_{1:k+1}$ and updating the model parameter $\bm{\theta}$ as well as the PCF $\alpha_{Z_{1:k}\rightarrow Z_{k+1}}$. We repeat this CPC process until the final stage $S_K$ is achieved. \begin{algorithm}
\caption{Training procedure of proposed CPC framework.} 
\label{alg3.1} 
\begin{algorithmic}
\REQUIRE observed sequences $\mathbf{X}_{1:T_{h}}$, ground truth future sequences $\mathbf{X}_{T_{h}+1:T_{h}+T_{p}}$, model parameters $\bm{\theta}$, stage number $K$, training epoch for $k$-th stage $E_{k}$, learning rate $\lambda$, training sample number $M$.  
\FOR{$i=1$ to $E_{1}$}
    \STATE $\mathbf{\hat{X}}_{T_{h}+1:T_{1}}=\mathbf{f}_{\bm{\theta}}(\mathbf{X}_{1:T_{h}})$
    \STATE $\bm{\theta} \leftarrow \bm{\theta} -\lambda*\nabla_{\bm{\theta}}\mathcal{L}_1(\mathbf{X}_{T_{h}+1:T_{1}}, \mathbf{\hat{X}}_{T_{h}+1:T_{1}})  $
\ENDFOR
\STATE $\mathbf{B}=\emptyset$
\FOR{$k=2$ to $K$} 
    \FOR{$i=1$ to $E_{k}$}
        \STATE $\mathbf{\hat{X}}_{T_{h}+1:T_{k}}, \alpha_{Z_{1:k-1}\rightarrow Z_{k}}=\mathbf{f}_{\bm{\theta}}(\mathbf{X}_{1:T_{h}})$
        \STATE $\bm{\theta} \leftarrow \bm{\theta} -\lambda*\nabla_{\bm{\theta}}\mathcal{L}_k(\mathbf{X}_{T_{h}+1:T_{k}}, \mathbf{\hat{X}}_{T_{h}+1:T_{k}}, \alpha_{Z_{1:k-1}\rightarrow Z_{k}}, \mathbf{B}) $
    \ENDFOR
    \STATE $\hat{\alpha}_{Z_{1:k-1}\rightarrow Z_{k}} = 0$
    \FOR{$m=1$ to $M$}
        \STATE $ \alpha^{m}_{Z_{1:k-1}\rightarrow Z_{k}} = \mathbf{f}_{\bm{\theta}}(\mathbf{X}^{m}_{1:T_{h}})$
        \STATE $\hat{\alpha}_{Z_{1:k-1}\rightarrow Z_{k}}= \hat{\alpha}_{Z_{1:k-1}\rightarrow Z_{k}} + \alpha^{m}_{Z_{1:k-1}\rightarrow Z_{k}}$
    \ENDFOR
    \STATE $\hat{\alpha}_{Z_{1:k-1}\rightarrow Z_{k}} = \frac{1}{M}\hat{\alpha}_{Z_{1:k-1}\rightarrow Z_{k}}$
    \STATE $\mathbf{B} = \mathbf{B} \cup \{\hat{\alpha}_{Z_{1:k-1}\rightarrow Z_{k}}\}$
\ENDFOR 
\end{algorithmic} 
\end{algorithm}

\section{Continual Prior Compensation++}
The Prior Compensation Factor in the CPC framework intends to encode the prior knowledge loss for all of the previous subtasks using a single random variable, which only provides a rough estimation of the prior knowledge loss. For example, when considering three tasks $Z_1, Z_2,$ and $Z_3$, if task $Z_{1}$ experiences significant prior information loss while task $Z_2$ does not, the PCF $\alpha_{Z_{1:2}\rightarrow Z_{3}}$ remains small due to the influence of task $Z_2$. Consequently, the $1-\alpha_{Z_{1:2}\rightarrow Z_{3}}$ will be large, preventing the model from effectively mitigating the prior forgetting problem of task $Z_1$. To handle this problem, we define the Fine-Grain Prior Compensation Factor (FGPCF), which contains multiple learnable random variables, each of which measures the prior information loss for the corresponding subtask. Based on the FGPCF, we formulate our Continual Prior Compensation++ (CPC++) framework, which intends to optimize the FGPCF and model parameters simultaneously in a multi-stage progressive training manner. We further derive a reasonable upper bound of the objective function for CPC++ through theoretical derivation, which can be efficiently optimized. In the following, we first provide a formal modeling and derivation of FGPCF as well as the optimization objective of CPC++ in Section \ref{subsec:4.1}. Section \ref{subsec:4.2} introduces the estimation of FGPCF. In Section \ref{subsec:4.3}, the optimization strategy is elaborated.

\subsection{Objective of Continual Prior Compensation++}
\label{subsec:4.1}
\subsubsection{\textbf{Fine-Grained Prior Compensation Factor}}
To estimate the prior information loss for each task, we propose the Fine-Grain Prior Compensation Factor (FGPCF), which is defined as follows:
\begin{equation}
\label{eq:4.1.1}
\alpha_{i}^{(k)}=\left\{
\begin{split}
    & P(Z_{2:k}|Z_{1};\bm{\theta}) - P(Z_{2:k}|\hat{Z}_{1};\bm{\theta}), \quad i=1,\\    
    & P(Z_{i+1:k}|\hat{Z}_{1:i-1}Z_{i};\bm{\theta}) - P(Z_{i+1:k}|\hat{Z}_{1:i};\bm{\theta}),\\
    & \quad \quad \quad \quad \quad \quad \quad \quad i\in\{2,3,\cdots, k-1\},
\end{split}
\right.
\end{equation}
where $k\in\{2,3,\cdots, K\}$. $P(Z_{i+1:k}|\hat{Z}_{1:i-1}Z_{i};\bm{\theta})$ indicates completing tasks $Z_{i+1:k}$ using corrupted prior information of tasks $Z_{1:i-1}$ and complete prior knowledge of task $Z_{i}$. And $P(Z_{i+1:k}|\hat{Z}_{1:i};\bm{\theta})$ represents completing tasks $Z_{i+1:k}$ with lost prior knowledge learned from tasks $Z_{1:i}$. Since the impact of prior information loss in previous tasks $Z_{1:i-1}$ is excluded using conditional probability, we can effectively measure the prior loss for the current task $Z_i$. Therefore, this approach enables a fine-grained assessment of information loss for each individual task, which leads to a more specific estimation of prior information loss and control of the training process.

\subsubsection{\textbf{Optimization Objective}}
Since the optimization function $-\ln P(Z_{1:k};\bm{\theta})$ is challenging to optimize directly, based on Lemma 3.3, we theoretically show that we can turn to minimize an upper bound of our optimization objective instead, which can be optimized easily. The upper bound is given as follows:
\begin{equation}
  \label{eq:4.3.2}
\begin{split}
    UB^{(k)} &= \sum_{i=2}^{k}(\prod \limits_{j=1}^{i-1} A_{j}^{(k)})(-\ln P(Z_{i}|\hat{Z}_{1:i-1};\bm{\theta})) \\
    &+ \sum_{i=2}^{k-1}(\prod \limits_{j=1}^{i-1} A_{j}^{(k)})\Delta_{i}^{(k)}  - \ln P(Z_{1};\bm{\theta}) + \Delta_{1}^{(k)},
\end{split}
\end{equation}
where $A_{i}^{(k)} = 1 - \alpha_{i}^{(k)},\quad \Delta_{i}^{(k)} = (1 - \alpha_{i}^{(k)})\ln(1 - \alpha_{i}^{(k)}) + \ln(1 + \alpha_{i}^{(k)}),\quad i \in\{1,2,\cdots, k-1\}$. In the following, we provide the description and proof of Lemma 3.3 in detail.

\textit{\textbf{Lemma 3.3.}} The inequality 
\begin{equation*}
\begin{split}
    -\ln P(Z_{1:k};\bm{\theta}) &\leq \sum_{i=2}^{k}(\prod \limits_{j=1}^{i-1} A_{j}^{(k)})(-\ln P(Z_{i}|\hat{Z}_{1:i-1};\bm{\theta})) \\
    &+ \sum_{i=2}^{k-1}(\prod \limits_{j=1}^{i-1} A_{j}^{(k)})\Delta_{i}^{(k)}  - \ln P(Z_{1};\bm{\theta}) + \Delta_{1}^{(k)}
\end{split}
\end{equation*}
holds. The equality holds if and only if $A_{i}^{(k)} = 1$ and $\Delta_{i}^{(k)} = 0$, $i\in\{1,2,\cdots, k-1\}$. 

\textit{Proof:} We can rewrite $-\ln P(Z_{1:k};\bm{\theta})$ using conditional formular:
\begin{equation}
\begin{split}
    -\ln P(Z_{1:k};\bm{\theta}) = -\ln P(Z_{2:k}|Z_{1};\bm{\theta}) - \ln P(Z_{1};\bm{\theta}).
\end{split}
\label{eq:ProofLe3.3_1}
\end{equation}
Using the definition from Equation (\ref{eq:4.1.1}) and Lemma 3.1, we can obtain:
\begin{equation}
\begin{split}
    -\ln P(&Z_{1:k};\bm{\theta}) = -\ln (P(Z_{2:k}|\hat{Z}_{1};\bm{\theta}) + \alpha_{1}^{(k)})  - \ln P(Z_{1};\bm{\theta}) \\
    &\leq  - \ln P(Z_{1};\bm{\theta}) + (1 - \alpha_{1}^{(k)})(-\ln (P(Z_{2:k}|\hat{Z}_{1};\bm{\theta})) \\
    & + (1-\alpha_{1}^{(k)})\ln(1-\alpha_{1}^{(k)}) + \ln(1+\alpha_{1}^{(k)}).
\end{split}
\label{eq:ProofLe3.3_2}
\end{equation}
For simplicity, let’s denote $A_{i}^{(k)} = 1 - \alpha_{i}^{(k)},\quad \Delta_{i}^{(k)} = (1 - \alpha_{i}^{(k)})\ln(1 - \alpha_{i}^{(k)}) + \ln(1 + \alpha_{i}^{(k)}),\quad i \in\{1,2,\cdots, k-1\}$. By repeatedly utilizing Equation (\ref{eq:4.1.1}) and Lemma 3.1, we can obtain:
\begin{equation}
\begin{split}
    & A_{1}^{(k)}(-\ln P(Z_{2:k}|\hat{Z}_{1};\bm{\theta})) + \Delta_{1}^{(k)} - \ln P(Z_{1};\bm{\theta})\\
    = & A_{1}^{(k)}(-\ln P(Z_{3:k}|\hat{Z}_{1}Z_{2};\bm{\theta}) -\ln P(Z_{2}|\hat{Z}_{1};\bm{\theta}) )  \\
    &+ \Delta_{1}^{(k)} - \ln P(Z_{1};\bm{\theta})\\
    = &  A_{1}^{(k)}(-\ln (P(Z_{3:k}|\hat{Z}_{1:2};\bm{\theta}) + \alpha_{2}^{(k)}))  + \Delta_{1}^{(k)} \\
    &+ A_{1}^{(k)}(-\ln P(Z_{2}|\hat{Z}_{1};\bm{\theta})) - \ln P(Z_{1};\bm{\theta})\\
    \leq &  A_{1}^{(k)}A_{2}^{(k)}(-\ln P(Z_{3:k}|\hat{Z}_{1:2};\bm{\theta})) + A_{1}^{(k)}\Delta_{2}^{(k)}  + \Delta_{1}^{(k)} \\
    &+ A_{1}^{(k)}(-\ln P(Z_{2}|\hat{Z}_{1};\bm{\theta})) - \ln P(Z_{1};\bm{\theta})\\
    \cdots \\
    \leq & \sum_{i=2}^{k}(\prod \limits_{j=1}^{i-1} A_{j}^{(k)})(-\ln P(Z_{i}|\hat{Z}_{1:i-1};\bm{\theta})) \\
    &+ \sum_{i=2}^{k-1}(\prod \limits_{j=1}^{i-1} A_{j}^{(k)})\Delta_{i}^{(k)}  - \ln P(Z_{1};\bm{\theta}) + \Delta_{1}^{(k)}.
\end{split}
\label{eq:ProofLe3.3_3}
\end{equation}
When $A_{i}^{(k)} = 1$ and $\Delta_{i}^{(k)} = 0$, $i\in\{1,2,\cdots, k-1\}$, the equality holds.
\begin{figure*}[ht]
\centering
{\includegraphics[width=\textwidth, height=0.2\textwidth]{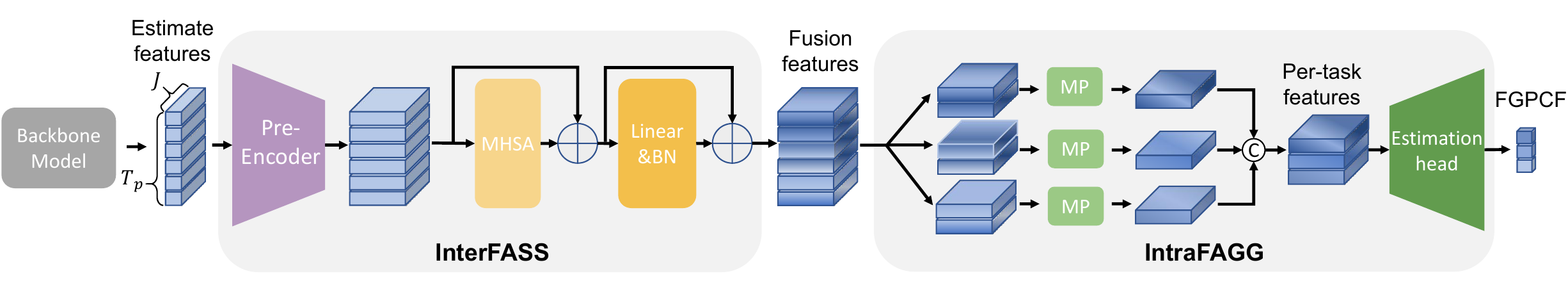}}
\hfil
\caption{Illustration of Prior Loss Estimation Module. InterFASS is the Inter-Task Forgetting Assessment module, and IntraFAGG represents the Intra-Task Forgetting Aggregation module. MP denotes the max-pooling layer.}
\label{fig:AEM}
\end{figure*}
\subsection{Estimation of FGPCF}
\label{subsec:4.2}
Here, we develop a Prior Loss Estimation Module (PLEM) to explicitly estimate the prior information loss. The PLEM consists of an Inter-Task Forgetting ASSessment (InterFASS) module and an Intra-Task Forgetting AGGregation (IntraFAGG) module, as illustrated in Figure \ref{fig:AEM}. Specifically, consider a HMP backbone model with an output of $\bm{M}_{res}\in \mathbb{R}^{T\times N\times 3}$\xt{, where $T$ and $N$ denotes the motion length and the number of joints, respectively}. We extend the model output with an additional dimension, generating an output of $\bm{M}_{res}\in \mathbb{R}^{T\times N\times (3+1)}$. We then extract features of the last dimension, with a size of $\bm{M}_{in}\in \mathbb{R}^{T\times N\times 1}$, and feed it into the InterFASS and IntraFAGG module, producing an output of $\bm{M}_{\alpha}\in \mathbb{R}^{K\times 1}$, which forms the estimation for FGPCF. The details of each module are elaborated in the following.

\noindent
\textbf{Inter-Task Forgetting Assessment Module.} Firstly, the model employs a pre-encoder to encode the input features into a high-dimensional space, facilitating the subsequent estimation of the forgetting extent of prior information for each task. Considering that the FGPCF to be calculated is closely related to all the earlier tasks, we employ a multi-head self-attention mechanism (MHSA) to interact with the features of each task. To extract the information forgotten at each time step, attention is applied to each time step among all tasks. Subsequently, the output of MHSA is fed into a linear mapping and batch normalization.

\noindent
\textbf{Intra-Task Forgetting Aggregation Module.} With the features after interaction through InterFASS provided, we then employ intra-task max-pooling to combine the features at each time step for various tasks. The combination features are then fed into an estimation head which contains a multilayer perceptron (MLP) and a sigmoid mapping. Finally, we can obtain an estimate of prior information loss (i.e., FGPCF) for each task on the given sample, which has a value from 0 to 1. 

\subsection{Optimization Strategy}
\label{subsec:4.3}
Similar to the CPC framework, CPC++ follows a multi-stage progressive training paradigm, in which a total of $K$ CPC++ stages are involved. In the first stage $S_1$, we aim to train the model for predicting the motion in the foremost segment without estimating FGPCF. While for the stage $S_k, k\in \{2,3,\cdots, K\}$, we update our model to predict motion segments for task $Z_{1:k}$ and simultaneously learn the FGPCF $\alpha_{i}^{(k)}, i\in\{1,2,\cdots, k-1\}$. This process is repeated until the final stage $S_K$ is achieved.

\noindent
\textbf{Learning of initial stage $S_1$.}
We train the initial stage $S_{1}$ with MSE loss:
\begin{equation}
  \label{eq:3.2.13}
  \begin{split}
    \mathcal{L}_1 
                &= \sum_{i=T_{h}+1}^{T_{1}} {\left\lVert \mathbf{X}_{i} - \hat{\mathbf{X}}_{i} \right\rVert}^2,
  \end{split}
\end{equation}
where $\mathbf{X}_{i}$ and $\hat{\mathbf{X}}_{i}$ represent the ground truth and predicted motion of the $i$-th frames respectively. 

\noindent
\textbf{Learning of stage $S_k$.}
In stage $S_{k}$ ($k \ge 2$), we need to update the model parameters $\bm{\theta}$ corresponding to tasks $Z_{1:k}$ and estimate the FGPCF $\alpha_{i}^{(k)}, i\in\{1,2,\cdots, k-1\}$. According to Equation (\ref{eq:4.3.2}), the loss function in this stage can be calculated as follows:
\begin{equation}
  \label{eq:3.2.15}
  \begin{split}
    \mathcal{L}_{k} &= \sum_{i=2}^{k}(\prod \limits_{j=1}^{i-1} A_{j}^{(k)})\sum_{v=T_{i-1}+1}^{T_{i}} {\left\lVert \mathbf{X}_{v} - \hat{\mathbf{X}}_{v} \right\rVert}^2   \\
    &+ \sum_{i=2}^{k-1}(\prod \limits_{j=1}^{i-1} A_{j}^{(k)})\Delta_{i}^{(k)} + \mathcal{L}_{1} + \Delta_{1}^{(k)}.
  \end{split}
\end{equation}
We continue the CPC++ training process by completing tasks $Z_{1:k+1}$ and updating the model parameter $\bm{\theta}$ as well as FGPCF $\alpha_{i}^{(k+1)}, i\in\{1,2,\cdots, k\}$. The algorithm flow is summarized in Algorithm \ref{alg4.1}, where the $\mathbf{B}^{(k)} = \{\alpha_{1}^{(k)}, \alpha_{2}^{(k)},\cdots,\alpha_{k-1}^{(k)}\}$.
\begin{algorithm}
\caption{Training procedure of proposed CPC++ framework.} 
\label{alg4.1} 
\begin{algorithmic}
\REQUIRE observed sequences $\mathbf{X}_{1:T_{h}}$, ground truth future sequences $\mathbf{X}_{T_{h}+1:T_{h}+T_{p}}$, model parameters $\bm{\theta}$, stage number $K$, training epoch for $k$-th stage $E_{k}$, learning rate $\lambda$.  
\FOR{$i=1$ to $E_{1}$}
    \STATE $\mathbf{\hat{X}}_{T_{h}+1:T_{1}}=\mathbf{f}_{\bm{\theta}}(\mathbf{X}_{1:T_{h}})$
    \STATE $\bm{\theta} \leftarrow \bm{\theta} -\lambda*\nabla_{\bm{\theta}}\mathcal{L}_1(\mathbf{X}_{T_{h}+1:T_{1}}, \mathbf{\hat{X}}_{T_{h}+1:T_{1}})  $
\ENDFOR
\FOR{$k=2$ to $K$} 
    \FOR{$i=1$ to $E_{k}$}
        \STATE $\mathbf{\hat{X}}_{T_{h}+1:T_{k}}, \mathbf{B}^{(k)}=\mathbf{f}_{\bm{\theta}}(\mathbf{X}_{1:T_{h}}),$
       \STATE $\bm{\theta} \leftarrow \bm{\theta}-\lambda*\nabla_{\bm{\theta}}\mathcal{L}_k(\mathbf{X}_{T_{h}+1:T_{k}}, \mathbf{\hat{X}}_{T_{h}+1:T_{k}}, \mathbf{B}^{(k)}) $
    \ENDFOR
\ENDFOR 
\end{algorithmic} 
\end{algorithm}
\section{Experiments}
\begin{table*}[ht]
\centering
  \caption{Results on Human3.6M, CMU-MoCap and 3DPW using PGBIG as baseline. A lower value means better performance.} 
 	\renewcommand{\arraystretch}{1.5}
 	\setlength{\tabcolsep}{2.2mm}
    \begin{tabular}{c|c|cccccc}
    \toprule
    \textbf{Dataset} & \textbf{Method} & 80ms  & 160ms & 320ms & 400ms & 560ms & 1000ms \\
    \midrule
    \multirow{3}[4]{*}{Human3.6M}  & PGBIG & 10.3   & 22.7  & 47.4  & 58.5  & 76.9  & 110.3 \\
    \cmidrule{2-8}                 & PGBIG\&CPC & 9.4 \textcolor{blue}{(-0.9)}   & 21.3 \textcolor{blue}{(-1.4)}  & 45.7 \textcolor{blue}{(-1.7)}  & 56.8 \textcolor{blue}{(-1.7)}  & 75.4 \textcolor{blue}{(-1.5)} & \textbf{108.8} \textcolor{blue}{(-1.5)} \\
    \cmidrule{2-8}                 & PGBIG\&CPC++ & \textbf{9.1} \textcolor{blue}{(-1.2)} & \textbf{20.9} \textcolor{blue}{(-1.8)}& \textbf{45.5} \textcolor{blue}{(-1.9)} & \textbf{56.6} \textcolor{blue}{(-1.9)} & \textbf{75.4} \textcolor{blue}{(-1.5)} & 109.2 \textcolor{blue}{(-1.1)}\\
    \midrule
    \midrule
    \multirow{3}[4]{*}{CMU-MoCap}  & PGBIG & 7.6   & 14.3  & 29.0  & 36.6  & 50.9  & 80.1  \\
    \cmidrule{2-8}                 & PGBIG\&CPC & \textbf{7.5 } \textcolor{blue}{(-0.1)} & 14.3 \textcolor{blue}{(-0.0)}  & 28.3 \textcolor{blue}{(-0.7)} & 35.4 \textcolor{blue}{(-1.2)} & 48.6 \textcolor{blue}{(-2.3)} & 78.4 \textcolor{blue}{(-1.7)} \\
    \cmidrule{2-8}                 & PGBIG\&CPC++ & 7.6 \textcolor{blue}{(-0.0)}  & \textbf{14.2 } \textcolor{blue}{(-0.1)} & \textbf{27.9 } \textcolor{blue}{(-1.1)} & \textbf{34.9 } \textcolor{blue}{(-1.7)} & \textbf{48.0 } \textcolor{blue}{(-2.9)}& \textbf{76.8 } \textcolor{blue}{(-3.3)}\\
    \midrule
    \midrule
    \multirow{3}[4]{*}{3DPW}  & PGBIG & 13.7  & 23.2  & 47.3  & 58.3  & 76.1  & 104.1  \\
    \cmidrule{2-8}            & PGBIG\&CPC & 7.2 \textcolor{blue}{(-6.6)}  & 16.3 \textcolor{blue}{(-6.9)} & 37.3 \textcolor{blue}{(-10.0)} & 47.1 \textcolor{blue}{(-11.2)} & \textbf{64.1} \textcolor{blue}{(-12.0)} & \textbf{96.0} \textcolor{blue}{(-8.1)}\\
    \cmidrule{2-8}            & PGBIG\&CPC++ & \textbf{7.1} \textcolor{blue}{(-6.7)} & \textbf{16.2} \textcolor{blue}{(-7.0)}& \textbf{37.1} \textcolor{blue}{(-10.2)} & \textbf{47.0} \textcolor{blue}{(-11.3)} & 64.5 \textcolor{blue}{(-11.6)} & 96.3 \textcolor{blue}{(-7.8)} \\
    \bottomrule
    \end{tabular}%
    \label{tab:4.2.0}%
\end{table*}
\subsection{Experimental Setup}%
\noindent
\textbf{Datasets.}
We validate our framework on three benchmark datasets: Human3.6M\cite{ionescu2013human3}, CMU-MoCap, 3DPW\cite{von2018recovering}. Human3.6M\cite{ionescu2013human3} is a large dataset containing 3.6 million 3D human pose data. 15 types of actions performed by 7 actors (S1, S5, S6, S7, S8, S9 and S11) are included in this dataset. Each actor is represented by a skeleton of 32 joints. However, following the data preprocessing method proposed in \cite{ma2022progressively,mao2019learning}, we only use 22 joints in the experiments. The global rotations and translations of poses are removed, and the frame rate is downsampled from 50 fps to 25 fps. We use actors S5 and S11 for testing and validation while conducting training on the remaining sections of the dataset. CMU-MoCap is a smaller dataset with 8 different action categories. The global rotations and translations of the poses are also removed. Following the data preprocessing methods in \cite{ma2022progressively,mao2019learning}, we use 25 joints to indicate human poses. 3DPW\cite{von2018recovering} is a challenging dataset that contains human motion data captured from both indoor and outdoor scenes. Poses in this dataset are represented in 3D space, with each pose containing 26 joints. However, only 23 of these joints are used, as the other three are redundant. 

\noindent
\textbf{Evaluation Metrics.} Following the benchmark protocols, we use the Mean Per Joint Position Error (MPJPE) in millimeters (ms) as our evaluation metric for 3D coordinate errors. We follow \cite{ma2022progressively,mao2019learning} to report both short-term (80, 160, 320 and 400ms) and long-term predictions (560 and 1000ms). The performance is better if this metric is smaller. 

\noindent
\textbf{Implementation Details.} Following \cite{ma2022progressively,mao2019learning,bouazizi2022motionmixer}, we set the input length to 10 frames and the predictive output to 25 frames for Human3.6M and CMU-Mocap datasets, respectively. For the 3DPW dataset, we predict 30 frames conditioned on the observation of the preceding 10 frames. We choose PGBIG \cite{ma2022progressively} as our baseline model by default. To learn the PCF, we add an extra dimension to the output of the backbone model and calculate PCF through an MLP network whose hidden dimension is set to 512. For estimating the FGPCF, we utilize the PLEM. The self-attention mechanism has four heads, and the hidden dimension of PLEM is configured as 64. We partitioned the future sequences into three segments with lengths of 3, 9, and 13. The training process was conducted on an NVIDIA RTX 3090 GPU for 120 epochs, allocating 50, 90, and 120 epochs for each stage. It should be noted that the inference process is the same as that of the baseline models.

\noindent
\textbf{Backbones.} We apply our method on the following backbone approaches: LTD \cite{mao2019learning}, MotionMixer (MM) \cite{bouazizi2022motionmixer}, siMLPe \cite{guo2023back} and the current state-of-the-art PGBIG \cite{ma2022progressively} and report the experimental results on three benchmark datasets. LTD and PGBIG are GCN-based models. MotionMixer and siMLPe are MLP-based models. All these methods have released their code publicly. We employ their pre-trained models or re-train their models using the suggested hyper-parameters for a fair comparison. And we also exactly follow the metric they used to evaluate the results.

\subsection{Main Experimental Results}
\noindent
\textbf{Quantitative results.} Table \ref{tab:4.2.0} presents experimental results of the baseline model PGBIG with and without our training strategy (CPC and CPC++) on the Human3.6M, CMU-MoCap and 3DPW datasets. As shown, our frameworks outperform the corresponding baseline model by a considerable margin across all three datasets. Specifically, we can observe that the improvement in long-term prediction (e.g., 1000 ms) is greater compared to short-term prediction (e.g., 80 ms), indicating that the prior knowledge provided by short-term prediction is more crucial for challenging long-term prediction tasks. The results on the Human3.6M dataset show that our CPC framework decreases the prediction error by an average of 1.5 (52.9 vs. 54.4). Additionally, we can obtain a 1.7 (52.7 vs 54.4) error reduction using CPC++. When using our CPC strategy to train the PGBIG \cite{ma2022progressively}, we obtain about a 1.0 (35.4 vs. 36.4 in terms of average MPJPE) performance improvement on the CMU-Mocap dataset. Furthermore, when the CPC++ strategy is employed, we can achieve an additional improvement of 0.5. It is worth noting that our framework outperforms PGBIG by a margin of 9.1 (44.7 vs. 53.8 in terms of average MPJPE) on the 3DPW dataset. We attribute this to our multi-stage training framework, which effectively utilizes more comprehensive prior knowledge for subsequent predictions in this challenging dataset with smaller sizes and more complex motions. However, it also makes the difficult for the model with more parameters to learn the more specific FGPCF using this challenging dataset, which leads to the comparable performance between CPC++ and CPC on the 3DPW dataset.

\begin{figure*}[!t]
\centering
    \includegraphics[width=0.96\textwidth]{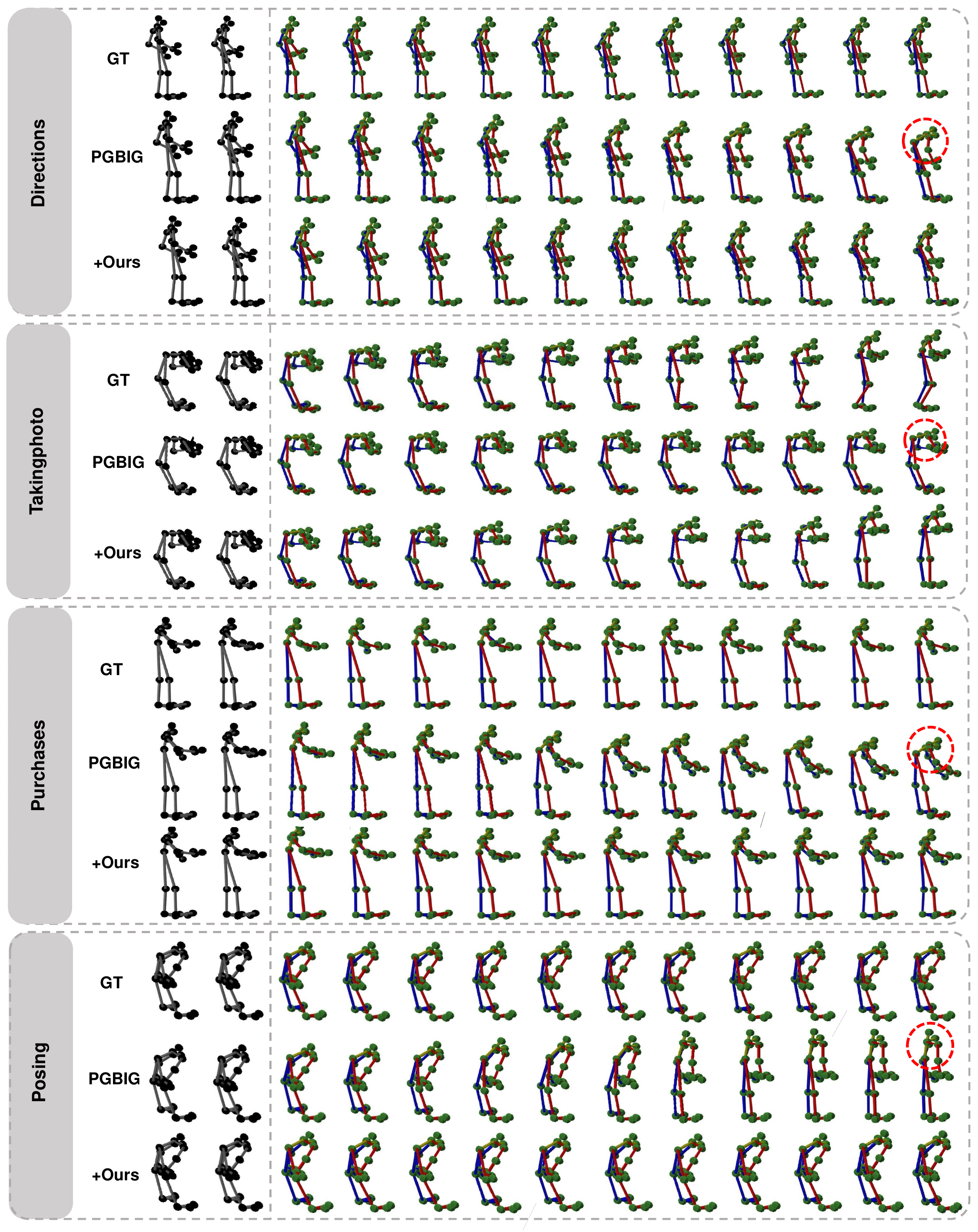}
\caption{Some visualized results from the Human3.6M dataset are presented. We display four action categories: ``Directions'', ``Takingphoto'', ``Purchases'', and ``Posing''. The observed frames are shown in black, while the colorful motion sequences represent the prediction results. ``GT'' denotes the ground truth results and ``PGBIG'' is the baseline. ``+Ours'' represents the results of using the proposed CPC framework. As demonstrated, our method produces the best future motion sequences. And we highlight the different parts in PGBIG's results with red dashed circles.}
  \label{fig.4.1}
\end{figure*}

\noindent
\textbf{Qualitative results.}
Figure \ref{fig.4.1} provides some visualization examples of predicted motions, qualitatively illustrating that our framework achieves more accurate results than the baseline model PGBIG. Specifically, in the ``Directions'' action, the person maintains an upright position throughout the sequence. In contrast, PGBIG exhibits a bent posture during long-term predictions, while our method accurately predicts positions closer to the ground truth. In the ``Takingphoto'' action, the person initially bends and then stands upright. PGBIG maintains the bent posture throughout long-term predictions, whereas our method accurately predicts posture changes. In the ``Purchases'' action, we observe that PGBIG's predicted results gradually bend, which differs from the ground truth. Furthermore, PGBIG's predicted results for the ``Posing'' action show an unnatural posture, where the person stands upright but with shoulders hunched forward and hands hanging vertically downward. It is evident that our method demonstrates an improvement in long-term prediction. This improvement is attributed to our proposed framework's ability to effectively leverage prior knowledge from short-term predictions, which provides long-term predictions with more comprehensive clues.
\begin{table*}[t]
\centering
  \caption{Results of other baseline models on Human3.6M. A lower value means better performance.}
 	\renewcommand{\arraystretch}{1.5}
 	\setlength{\tabcolsep}{2.6mm}
    \begin{tabular}{c|c|cccccc}
    \toprule
    \textbf{Method} & \textbf{Framework}  & 80ms  & 160ms & 320ms & 400ms & 560ms & 1000ms \\
    \midrule
    \multirow{3}[4]{*}{LTD}  &  baseline   & 12.7  & 26.1  & 52.3  & 63.5  & 81.6  & 114.3  \\
    \cmidrule{2-8}          &  CPC & 10.8 \textcolor{blue}{(-1.9)}  & 23.0 \textcolor{blue}{(-3.1)}  & 48.2 \textcolor{blue}{(-4.1)}  & 59.3 \textcolor{blue}{(-4.2)}  & 77.5 \textcolor{blue}{(-4.1)} & 111.2 \textcolor{blue}{(-3.1)} \\
                            & CPC++ & \textbf{10.2} \textcolor{blue}{(-2.5)} & \textbf{22.3} \textcolor{blue}{(-3.8)} & \textbf{47.3} \textcolor{blue}{(-5.0)} & \textbf{58.5} \textcolor{blue}{(-5.0)} & \textbf{77.2} \textcolor{blue}{(-4.4)} & \textbf{111.0} \textcolor{blue}{(-3.3)} \\
    \midrule
    \midrule
    \multirow{3}[4]{*}{MM}  &  baseline & 12.7  & 26.4  & 53.4  & 65.0  & 83.6  & 117.6  \\
    \cmidrule{2-8}          &  CPC & 11.5 \textcolor{blue}{(-1.2)} & 25.0 \textcolor{blue}{(-1.4)} & 52.0 \textcolor{blue}{(-1.4)}  & 63.7 \textcolor{blue}{(-1.3)} & 82.8 \textcolor{blue}{(-0.8)} & 117.1 \textcolor{blue}{(-0.5)} \\
                            & CPC++ & \textbf{10.8} \textcolor{blue}{(-1.9)} & \textbf{24.3} \textcolor{blue}{(-2.1)} & \textbf{51.4} \textcolor{blue}{(-2.0)} & \textbf{63.1} \textcolor{blue}{(-1.9)} & \textbf{82.4} \textcolor{blue}{(-1.2)} & \textbf{117.0} \textcolor{blue}{(-0.6)} \\
    \midrule
    \midrule
    \multirow{3}[4]{*}{siMLPe} & baseline  & 10.7  & 23.9  & 50.7  & 62.6  & 82.0  & 116.0  \\
    \cmidrule{2-8}          & CPC & 10.0 \textcolor{blue}{(-0.7)} & 22.9 \textcolor{blue}{(-1.0)} & 49.4 \textcolor{blue}{(-1.3)} & 61.2 \textcolor{blue}{(-1.4)} & 80.6 \textcolor{blue}{(-1.4)} & 114.6 \textcolor{blue}{(-1.4)} \\
                            & CPC++ & \textbf{10.0} \textcolor{blue}{(-0.7)} & \textbf{22.8} \textcolor{blue}{(-1.1)} & \textbf{48.9} \textcolor{blue}{(-1.8)} & \textbf{60.6} \textcolor{blue}{(-2.0)} & \textbf{80.2} \textcolor{blue}{(-1.8)} & \textbf{114.2} \textcolor{blue}{(-1.8)} \\
    \bottomrule
    \end{tabular}%
  \label{tab:4.2.1}%
\end{table*}

\begin{figure*}[ht]
  \centering
  \includegraphics[width=0.95\textwidth]{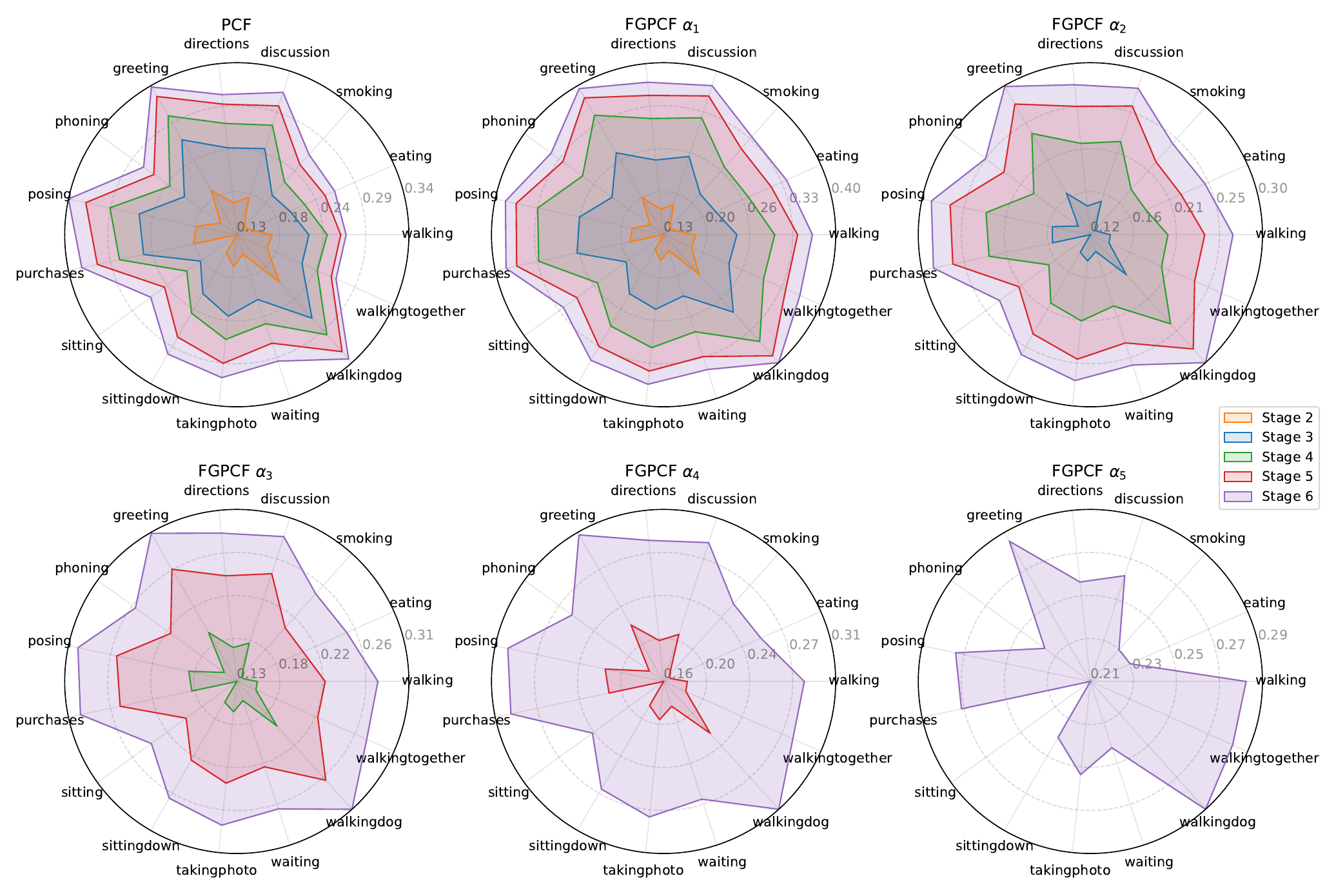}
  \caption{Visualized results of PCF and FGPCF. Each subplot is represented in polar coordinates, where the angular coordinate represents different action categories, and the radial distance represents values. The drawn circular area represents the results of a certain stage.}
  \label{fig:5.vis_alpha}
\end{figure*}

\noindent
\textbf{Visualization of PCF and FGPCF.} In our CPC and CPC++ frameworks, we have employed the PCF and FGPCF to mitigate the loss of prior knowledge. In Figure \ref{fig:5.vis_alpha}, we show the value of them in different stages and different action types.

The PCF value progressively increases with each training stage for all action types. However, the trends in change for different actions are not uniform. For example, the value of ``greeting'' is smaller than that of ``walkingdog'' at the first stage but becomes larger at stage 5. This difference can be attributed to the static nature of the ``greeting'' action in the short term, where the action pattern undergoes minimal change. Consequently, the forgetting problem of prior knowledge learned in the initial stages is less severe, leading to a smaller PCF value. Conversely, the ``greeting'' action manifests in various forms over long-term sequences, resulting in different action patterns and a more severe forgetting problem. Additionally, we observe that the PCF value of ``walking'' is higher than that of ``waiting'' at stage 1 but lower at stage 5. This suggests that ``walking'' exhibits more movement in short-term sequences. As a result, when the periodic action pattern is learned in the earlier stages, the loss of prior knowledge is less pronounced in stage 5. In contrast, ``waiting'' remains static initially but undergoes posture changes in long-term sequences. \xt{By exactly examining the PCF/FGPCF results in Figure \ref{fig:5.vis_alpha}, we find that the learned FGPCF values increase with stage progresses for all action types, which means that the prior-information loss induced by multi-stage continual training becomes more severe. These results confirm the effectiveness of our approach in learning fine-grained prior compensation factors for mitigating the prior information forgetting in the multi-stage continual training. Furthermore, we can also observe that both the PCF and FGPCF values vary across actions. In particular, low-dynamic actions (e.g., sitting, smoking) exhibit consistently smaller PCF/FGPCF values. In contrast, for the actions with diverse and complex dynamics (e.g., walking dog and greeting), we observe that the learned PCF/FGPCF values are much larger in all stages. These observations suggest that exploring dynamic-specific prior information compensation mechanisms for motion prediction, by explicitly considering dynamic similarities across actions, could be beneficial and promising. 
}
 

\subsection{Ablation Studies}
We conduct several ablation experiments to further verify the effectiveness of our proposed framework. 

\noindent
\textbf{Integration with different baseline models.}
We first study the flexibility by applying the CPC and CPC++ frameworks to progressively train different baseline models. Table \ref{tab:4.2.1} shows the quantitative comparisons of prediction results of different baseline models on the Human3.6M dataset. As shown, the performance of both short-term and long-term prediction improves when employing CPC and CPC++ in these baseline models. To be specific, we achieve a 3.6 performance improvement on average by applying CPC to LTD and a 4.1 performance improvement when using CPC++. Likewise, for the baseline models MM, CPC leads to a 1.1 performance enhancement, while CPC++ leads to a 1.6 performance enhancement. As for siMLPe, CPC also enhances prediction accuracy by approximately 1.2 and CPC++ improves the system performance by 1.5. These experimental results demonstrate that our proposed frameworks can effectively leverage prior knowledge to enhance the performance of both short and long-term prediction. Furthermore, these also validate the flexibility of applying CPC and CPC++ to various HMP baseline models, enhancing their performance.

\noindent
\textbf{Effect of PCF and FGPCF.}
In this paper, we have introduced the prior compensation factor and fine-grained prior compensation factor to mitigate the prior forgetting problem during the multi-stage training process. Here, we investigate the benefits of them. In our experiments, we first report the results of the baseline ``Without PCF'', which trains in a multi-stage manner without using the PCF. We also report the results of using PCF ``With PCF'' and using FGPCF ``With FGPCF''. \begin{figure}[t]
  \centering
  \includegraphics[width=0.45\textwidth]{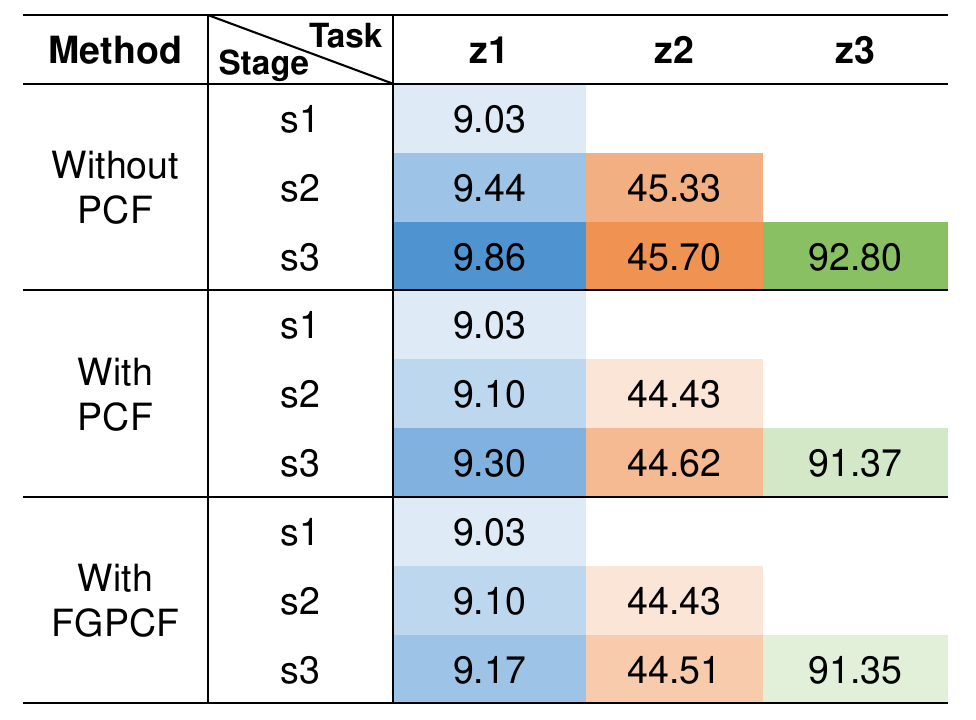}
  \caption{The average error of different tasks at the end of each stage. A lower value means better performance.}
  \label{fig:abla_pcf_fgpcf}
\end{figure}

The results presented in Figure \ref{fig:abla_pcf_fgpcf} show that introducing the PCF alleviates the performance degradation from stage $S_1$ to stage $S_3$ of task $Z_1$'s predictions. Specifically, without PCF, the prediction error of task $Z_1$ increases by 0.83, whereas with PCF, it only increases by 0.27. This result suggests that PCF can effectively alleviate the prior forgetting issue. As a result, task $Z_1$ can offer more comprehensive priors for tasks $Z_2$ and $Z_3$, leading to better prediction performance. To be specific,  the prediction error for task $Z_2$ at the end of stage $S_2$ is 45.33 without PCF, but it reduces to 44.43 when PCF is used. Similarly, for task $Z_3$ in stage $S_3$, the error decreases from 92.80 to 91.37 by using PCF. Furthermore, the results in Figure \ref{fig:abla_pcf_fgpcf} demonstrate that FGPCF further mitigates the loss of prior knowledge. The prediction error of task $Z_1$ only increases by 0.14, significantly smaller than that of the baseline model. Therefore, we obtain better performance of tasks $Z_2$ and $Z_3$ by utilizing more comprehensive prior knowledge. These promising experimental results confirm that the proposed PCF and FGPCF help mitigate the prior forgetting problem. 

\noindent
\textbf{Evaluation on Prior Loss Estimation Module.}
In our CPC++ framework, we have proposed the prior loss estimation module to better estimate the FGPCF, which contains the inter-task forgetting assessment and intra-task forgetting aggregation modules. Here, we conduct experiments to validate the effectiveness of these modules. In our experiments, we first remove both modules and obtain the baseline, ``w/o both''. Then, the result of adding InterFASS is reported as ``with InterFASS''. Furthermore, we test the result by only adding IntraFAGG, labeled as ``with IntraFAGG''. Finally, we use both modules, which is referred to as ``with both''. \begin{table}[htbp]
  \centering
  \renewcommand{\arraystretch}{1.3}
 	\setlength{\tabcolsep}{1.9mm}
  \caption{Results of our method with/without InterFASS and IntraFAGG modules.}
    \begin{tabular}{c|cc|c}
    \hline
    Method & InterFASS & IntraFAGG & Avg$\downarrow$ \\
    \hline
    w/o both &       &       &  65.7 \\
    with InterFASS & \checkmark &       &  65.0 \\
    with IntraFAGG &       & \checkmark &  65.5 \\
    with both & \checkmark & \checkmark &  \textbf{64.9} \\
    \hline
    \end{tabular}%
  \label{tab:abla_aem}%
\end{table}%

The results are presented in Table \ref{tab:abla_aem}. We can observe that without the InterFASS and IntraFAGG, the prediction error is 65.7. When InterFASS is employed, the error decreases to 65.0, demonstrating the importance of accurately estimating the extent of prior knowledge loss. The prediction error is reduced to 65.5 by using IntraFAGG, which indicates that IntraFAGG is beneficial for the estimation of prior knowledge loss. With both modules added, the error further drops to 64.9.

\noindent
\textbf{Evaluation on different implementations.}
In our CPC framework, we introduce the prior compensation factor to address the issue of prior forgetting. This factor is adaptively determined based on the backbone output. To demonstrate the importance of incorporating PCF, learned during the training process, we have conducted a series of experiments. In our experiments, we compare the results of four different implementations. PGBIG is the baseline model. ``w/o $\alpha$'' means that we only divide the training process into several stages to train each task without using the PCF. ``HC'' represents using a hand-crafted coefficient that changes its values similar to our PCF at each epoch. Specifically, in stage $S_1$, the value of $\alpha$ is set to 1. In stage $S_2$, $\alpha$ is initially set to 0.1 and increased by 0.05 at each epoch until reaching 0.5, where it remains constant. The same pattern applies to stage $S_3$. \begin{figure*}[ht]
  \centering
  \includegraphics[width=0.98\textwidth]{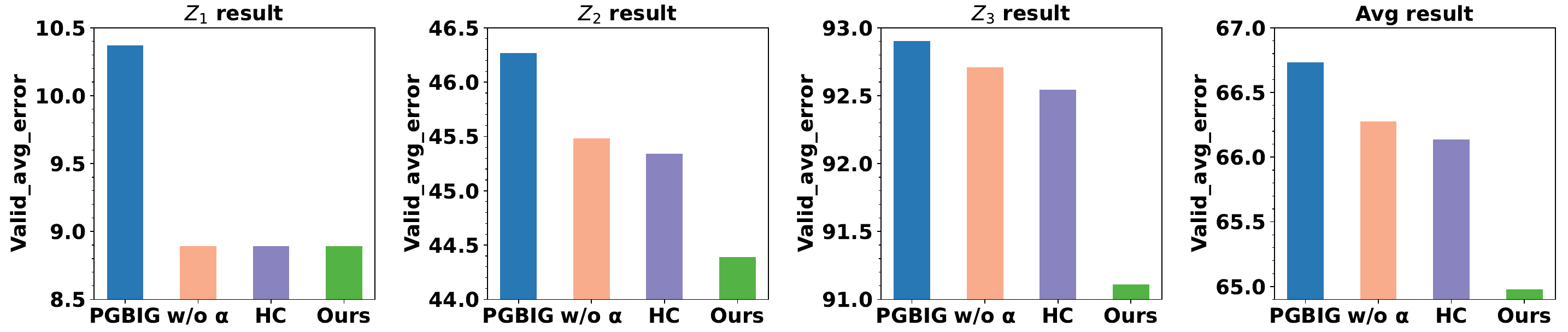}
  \caption{Comparison of different approaches. PGBIG is a baseline model that is trained without multi-stage. ``w/o $\alpha$'' represents training multi-stage process without PCF. ``HC'' means using a manually designed coefficient. ``Ours'' is a multi-stage training process with PCF. A lower value means better performance.}
  \label{fig.4.abla_loss_knowl}
\end{figure*}

The results are presented in Figure \ref{fig.4.abla_loss_knowl}. In each subfigure, we show the results of task $Z_k$ on the validation set which is obtained at stage $S_k$. In stage $S_1$, where no prior information is available, the results of "w/o $\alpha$", "HC", and "Ours" are identical, but superior to the baseline. This validates the effectiveness of decomposing multiple-moment predictions, as it alleviates the constraint of long-term predictions on short-term predictions and enhances the model's ability to learn short-term predictions. In the following stages, the performance of ``w/o $\alpha$'' is worse than ``Ours'', which indicates that the prior knowledge exploited by our framework benefits the prediction model training. Moreover, as the training period progressed, the performance gap became larger. However, the method without PCF can still achieve better performance than ``PGBIG'', indicating that the training model in a multi-stage manner can also exploit some useful prior information for prediction. We also note that the performance of ``Ours'' is much better than ``HC'', which conducts temporal continual learning with fixed and manually defined PCF. It demonstrates that joint training PCF and model parameters is beneficial.

\noindent
\textbf{Evaluation on the number of tasks.}
In most of our implementations, our CPC framework divides the future prediction into three tasks. Here, we study the influence of the number of tasks. Specifically, 1 task represents the training of short and long-term prediction together. In the 2 tasks setting, we partition the future sequences into two segments with lengths of 3 and 22. As for 5 tasks, we design prediction lengths of 3, 4, 5, 6 and 7 for each task. We split the future sequences into eight segments with lengths of 3, 3, 3, 3, 3, 3, 3 and 4 for the 8-tasks setting. The detailed comparison results are shown in Table \ref{tab:4.4.1}. As shown, the model's performance improves as the number of tasks gets larger from 1 to 3, and it remains stable when the number of tasks becomes larger than 3. It demonstrates that the proposed framework for task division is beneficial for human motion prediction. \begin{table}[h]
\centering
\renewcommand{\arraystretch}{1.5}
\setlength{\tabcolsep}{1.5mm}
\caption{The average error of different numbers of tasks.}
    \begin{tabular}{c|ccccc}
    \hline
    number of tasks & 1     & 2     & 3     & 5     & 8 \\
    \hline
    avg error$\downarrow$ & 66.95 & 66.02 & \textbf{65.00}    & 65.05 & 65.03 \\
    \hline
    \end{tabular}%
    \label{tab:4.4.1}
\end{table}

\xt{
\textbf{Effect of the overlapping vs. non-overlapping segments.}
In our progressive training framework, we split motions into non-overlapping segments and perform temporal continual learning across segments with prior compensation. To verify the effect of this design, we also implemented our approach using overlapping segments. The detailed results are presented in Table \ref{tab:overlap}. As expected, non-overlapping training yields the best performance, significantly outperforming implementations with overlapping segments, demonstrating the effectiveness of our proposed continual learning framework with prior compensation.
}

\begin{table}[t]
\centering
\renewcommand{\arraystretch}{1.5}
\setlength{\tabcolsep}{1.5mm}
\caption{Comparison results (MPJPE) of our method with different overlaps. X-overlap denotes the X-frame overlap between adjacent segments.}
    \begin{tabular}{c|cccccc}
    \hline
    Method & 80ms  & 160ms & 320ms & 400ms & 560ms & 1000ms \\
    \hline
    PGBIG (Baseline) & 10.3 & 22.7 & 47.4 & 58.5 & 76.9 & 110.3 \\
    6-overlap & 9.9 & 21.9 & 46.5 & 57.6 & 76.5 & 110.4 \\
    3-overlap & 9.8 & 21.8 & 46.5 & 57.6 & 76.3 & 110.3 \\
    no-overlap & \textbf{9.1} & \textbf{20.9} & \textbf{45.5} & \textbf{56.6} & \textbf{75.4} & \textbf{109.2} \\
    \hline
    \end{tabular}%
    \label{tab:overlap}
\end{table}

\xt{
\textbf{Evalution on the single PCF/FGPCF prediction head across stages.}
In our framework, we develop a single shared head to predict PCF/FGPCF across stages. Note that employing K output heads to generate K sub-predictions is also a feasible way to implement multi-stage training, in which each head mainly handles the prediction of a certain stage. However, this design faces scalability issues in long-term motion prediction since the number of head would grow with the total motion length T (e.g., K=[T/$\Delta$], where $\Delta$ represents length of motion predicted by each head), causing parameter bloat, higher memory/compute, and training instability. Moreover, the predictions outputted by multi-head decoder could encounter a termporal consistency issue, especially when the snippts are not overlapped. To emprically verify the effectivess of our single shared head, we also implement our multi-stage training with multi-head decoders and combined it with SOTA model PGBIG. The detailed results on Human3.6M dataset are presented in Table \ref{tab:K-head}. As shown, our method with single shared-head obtains the best prediction performances and outperforms the multi-head variant in both short-term and long-term motion prediction, demonstrating the effectiveness of our approach in leveraging short-term priors to progressively train models across stages. We can also observe that performing multi-stage training with multi-head decoder improves short-term prediction (80–400ms) over PGBIG but slightly degrades long-term prediction (500–1000ms) accuracy, indicating that short-term priors are not effectively exploited for long-term motion prediction in this design.
}

\begin{table}[t]
\centering
\renewcommand{\arraystretch}{1.5}
\setlength{\tabcolsep}{1.5mm}
\caption{Comparison results (MPJPE) of implementing multi-stage training with different strategies (single head vs. multi-head).}
    \begin{tabular}{c|cccccc}
    \hline
    Method & 80ms  & 160ms & 320ms & 400ms & 560ms & 1000ms \\
    \hline
    PGBIG & 10.3 & 22.7 & 47.4 & 58.5 & 76.9 & 110.3 \\
    \makecell{+multi-head} & 10.0 & 22.0 & 47.0 & 58.0 & 77.1 & 111.1 \\
    \makecell{+single shared-head} & 9.4 & 21.3 & 45.7 & 56.8 & 75.4 & 108.8 \\
    \hline
    \end{tabular}%
    \label{tab:K-head}
\end{table}

\xt{
\textbf{Analysis on the training time.}
In Table \ref{tab:training_time}, we provide the detailed training time of different methods with our progressive training strategy. As shown, our three-stage training strategy only slightly increases the total training time (e.g., from 16.3h to 19.5h for the PGBIG backbone) while significantly improving the overall performance from 64.9 to 67.0. The slight increase in training time can be attributed to the fact that learning the short-term prediction is much easier than optimizing over the entire sequence (8.1h vs. 16.3h), while the training in the subsequent stages can converge faster by leveraging the prior knowledge pretrained in previous stages. It is worth noting that our method is only used in the training stage, the inference time remains identical to that of the baseline model.
}
\begin{table}[h]
\centering
\renewcommand{\arraystretch}{1.5}
\setlength{\tabcolsep}{1.5mm}
\caption{Total training time comparison. The training procedure is stopped when the loss variation remains below 1e-5 within 5 epochs.}
    \begin{tabular}{c|ccccc}
    \hline
    Method & Avg. err.  & stage 1 & stage 2 & stage 3 & total \\
    \hline
    PGBIG & 67.0 & 16.3 h & - & - & 16.3 h \\
    PGBIG+Ours & 64.9 & 8.1 h & 6.0 h & 5.4 h & 19.5 h \\
    \hline\hline
    LTD & 70.2 & 8.4 h & - & - & 8.4 h \\
    LTD+Ours & 67.3 & 5.3 h & 3.0 h & 3.1 h & 11.4 h \\
    \hline
    \end{tabular}%
    \label{tab:training_time}
\end{table}

\begin{figure}[t]
  \centering
  \includegraphics[width=\linewidth]{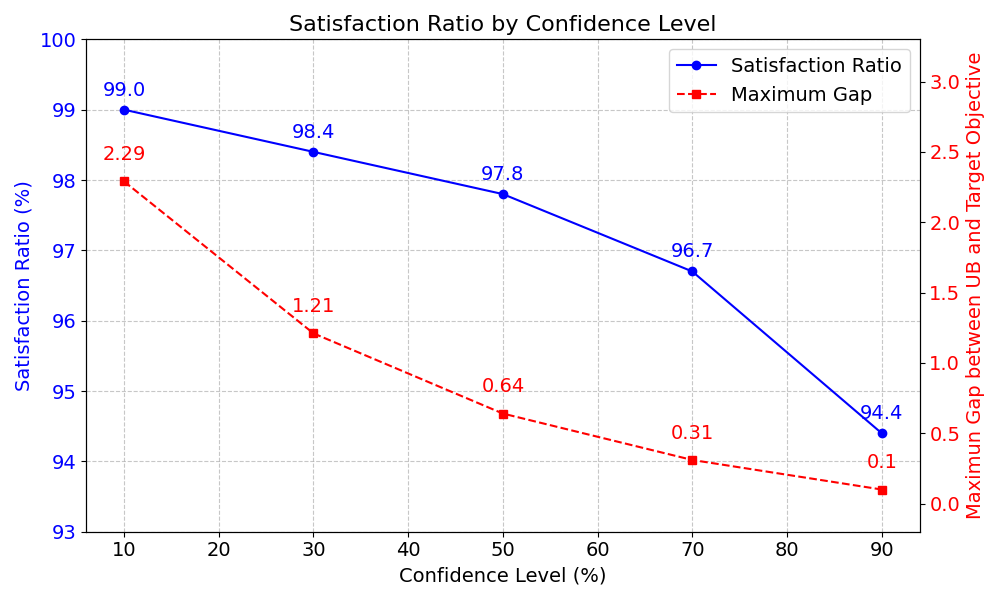}
  \caption{The ratio of samples and the upper bound of maximun gap between UB and target objective when the confidence value is larger than a certain level.}
  \label{fig.prob0.5}
\end{figure}

\xt{
\textbf{Discussion on the assumption $P(Z_i|\hat{Z}_{1:i-1};\theta) \geq \nicefrac{1}{2}$.}
In \textbf{Lemma 3.2}, we have derived the maximum difference between target objective and the upper bound (UB) under the assumption $P(Z_i|\hat{Z}_{1:i-1};\theta) \geq \nicefrac{1}{2}$. To validate the plausibility of this assumption, we have carefully examined the samples in our experiments on the Human3.6M dataset with PGBIG as baseline, and present Figure \ref{fig.prob0.5} to demonstrate the ratio of samples and maximum gap between UB and target objective when the confidence value is larger than a certain level. We empirically find that a relatively large number of samples can satisfy the condition/assumption. For instance, 97.8\% of the samples satisfy the condition $P(Z_i| \hat{Z}_{1:i-1}; \theta)>=\nicefrac{1}{2}$. When the confidence level decreases, the number of samples meeting the condition increases, while the maximum gap between UB and target objective becomes larger accordingly. For non-satisfying cases, we cannot mathematically derive an exact formulation for the maximum gap between UB and target objective, but we empirically examine the non-satisfying samples and find that their gaps are still in a reasonable range (from $0.64$ to $2.29$). And more importantly, such samples account for only $2.2\%$ of the total. 
}

\xt{
\textbf{Discussion on the training strategy of our method and Transformer.} Both our method and the Transformer involve processing sequences of varying lengths during training. Specifically, transformer processes all the sequences of varied lengths simultaneously in a single stage. In contrast, our method follows multi-stage training strategy, which processes the sequences of a certain length in each training stage. Here, we conduct experiments on the Human3.6M dataset and compare our results with POTR \cite{POTR}, which is trained following the standard Transformer paradigm. To ensure a fair comparison, we use the same EAE metric (in radians, ranging from 0 to $\pi$) as POTR for evaluation. The detailed results are presented in Table \ref{lab:trainingstr}. As shown, applying our temporal continual training strategy to POTR achieves better performance than the standard Transformer paradigm and the performance gain is much larger for long-term motion prediction (e.g., 400ms-1000ms) than short-term motion prediction (e.g., 80ms-160ms), demonstrating the effectiveness of our progressive multi-stage training paradigm to explore prior knowledge for facilitating human motion prediction. 
}

\begin{table}[t]
\centering
\renewcommand{\arraystretch}{1.5}
\setlength{\tabcolsep}{1.5mm}
\caption{Experimental results on Transformer-based model POTR with different training strategies.}
    \begin{tabular}{c|cccccc}
    \hline
    Method & 80ms  & 160ms & 320ms & 400ms & 560ms & 1000ms \\
    \hline
    Transformer & 0.235         & 0.581          & 0.990          & 1.143          & 1.362          & 1.826 \\
    \makecell{Ours} & 0.230  & 0.553  & 0.921  & 1.051   & 1.266 & 1.729 \\
    \makecell{Gain} & 0.005	&0.028 	&0.069	&0.092 &	0.096 &	0.097 \\    
    \hline
    \end{tabular}%
    \label{lab:trainingstr}
\end{table}

\section{Conclusion}
In this work, we proposed to progressively train the human motion prediction model in the temporal continual learning framework. Specifically, two different multi-stage training approaches (i.e., continual prior compensation and continual prior compensation++) are developed to progressively train the human motion prediction model. We introduce the prior compensation factor and fine-grained prior compensation factor to explicitly mitigate the information-forgetting problem that occurs in multi-stage model training. Furthermore, we theoretically show that the PCF and FGPCF can be efficiently learned together with the model parameters by minimizing a reasonable upper bound of the objective function. Extensive experiments demonstrated our framework's effectiveness and flexibility. We believe our work has value for not only human motion prediction but also for more general prediction tasks and backbone models \cite{sun2019predicting,lin2021predictive,hu2021apanet,mohamed2020social}, which forms one of our future work.

\section*{Acknowledgments}
This work was supported by the National Natural Science Foundation of China (Grant No. 62476296), Guangdong Natural Science Funds Project (2023B1515040025, 2022B1111010002), Guangdong NSF for Distinguished Young Scholar (2022B1515020009), Guangdong Provincial Key Laboratoryof Information Security Technology (2023B1212060026).



\section*{References}

\bibliographystyle{IEEEtran}
\bibliography{reference}
\section*{Biography Section}
\begin{IEEEbiography}[{\includegraphics[width=1in,height=1.25in,clip,keepaspectratio]{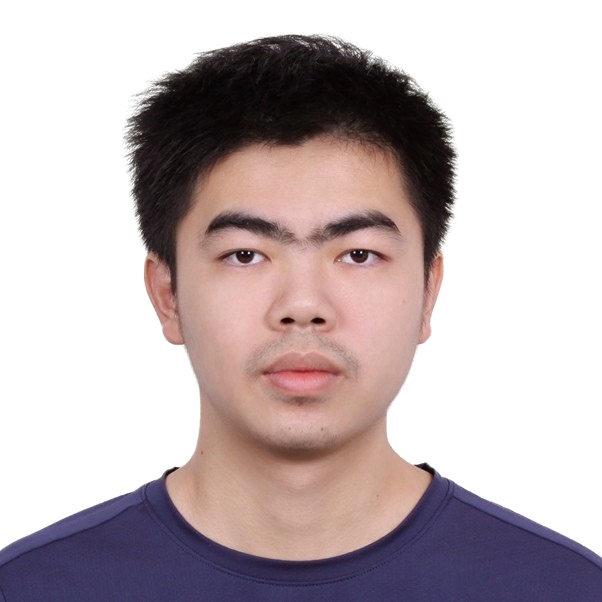}}]
{Jianwei Tang} received the bachelor's degree in computer science from Sun Yat-sen University in 2022. He is now a M.S. student in the School of Computer Science and Engineering in Sun Yat-sen University. His research interests include 3D human motion analysis and continual learning.
\end{IEEEbiography}
\begin{IEEEbiography}[{\includegraphics[width=1in,height=1.25in,clip,keepaspectratio]{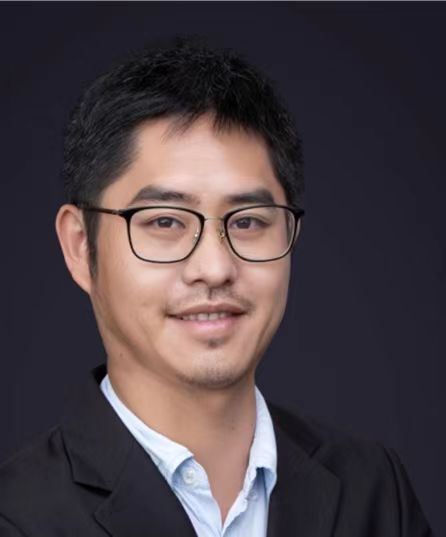}}]
{Jian-Fang Hu} is now an associate professor with Sun Yat-sen University. He received the Ph.D. and B.S. degrees from the School of Mathematics, Sun Yat-Sen University, Guangzhou, China, in 2016 and 2010, respectively. His research interests include human-object interaction modeling, 3D face modeling, and RGB-D action recognition. He has published several scientific papers in the international conferences and journals including ICCV, CVPR, NeurIPS, ECCV, and IEEE TPAMI. 
\end{IEEEbiography}
\begin{IEEEbiography}[{\includegraphics[width=1in,height=1.25in,clip,keepaspectratio]{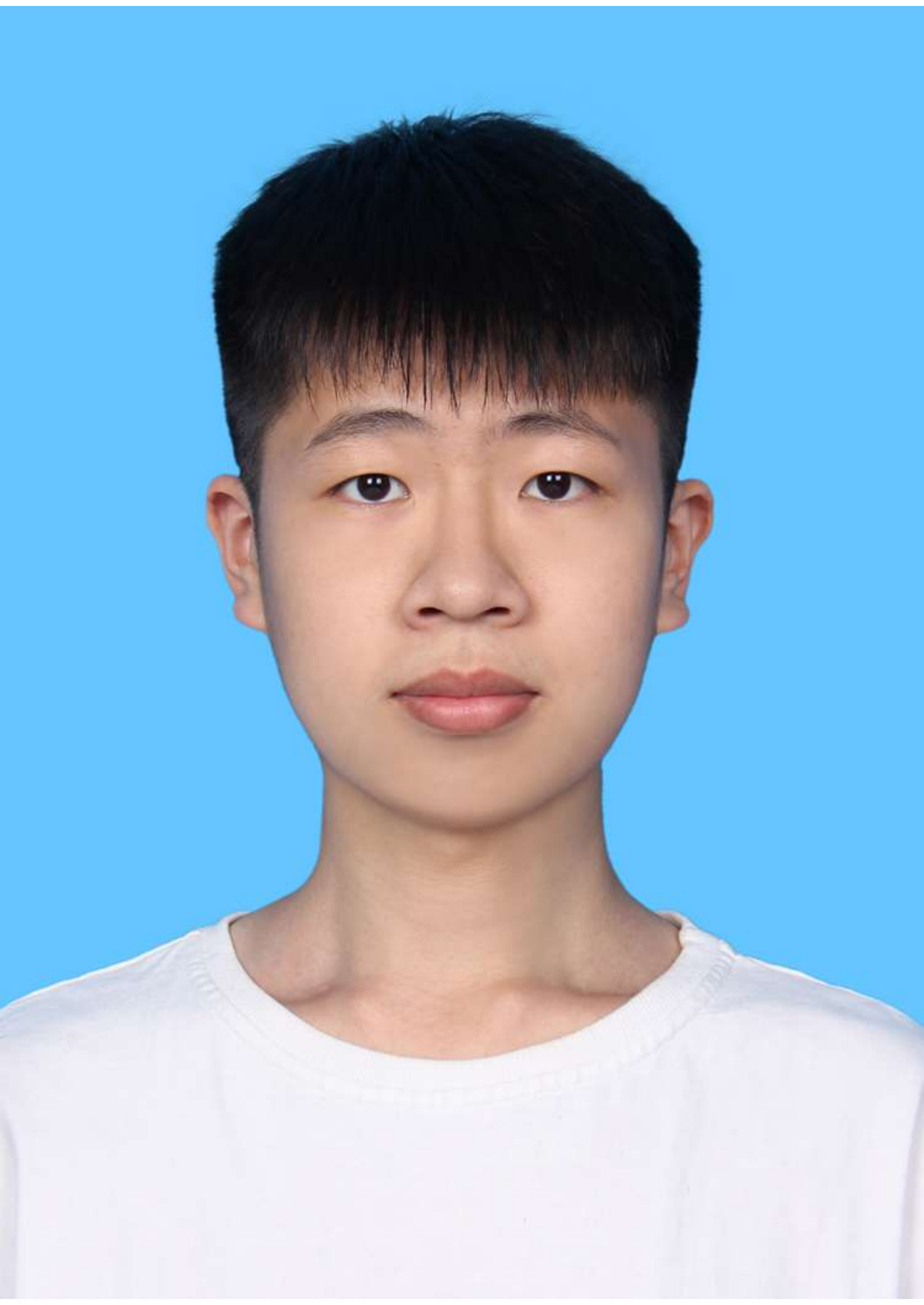}}]
{Tianming Liang} received the B.S. degree from Dalian University of Technology (DUT), and the M.S. degree from Harbin Institute of Technology (HIT). He is currently working toward the Ph.D. degree with the School of Computer Science and Engineering, Sun Yat-sen University, Guangzhou, China. His research interests include video understanding and vision-language learning.
\end{IEEEbiography}
\begin{IEEEbiography}[{\includegraphics[width=1in,height=1.25in,clip,keepaspectratio]{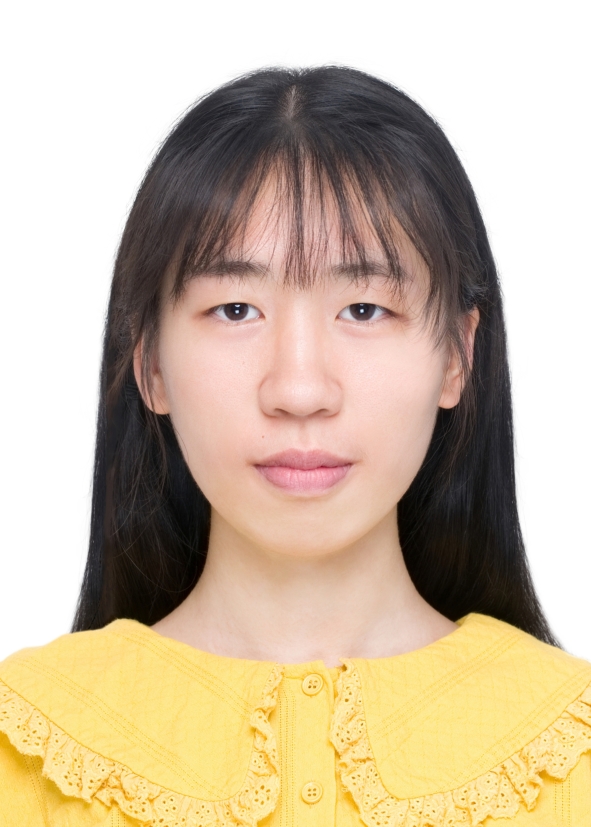}}]
{Xiaotong Lin} received the bachelor's degree in computer science from Sun Yat-sen University in 2022. She is now a M.S. student in the School of Computer Science and Engineering in Sun Yat-sen University. Her research interests include trajectory prediction and 3D human motion.
\end{IEEEbiography}
\begin{IEEEbiography}[{\includegraphics[width=1in,height=1.25in,clip,keepaspectratio]{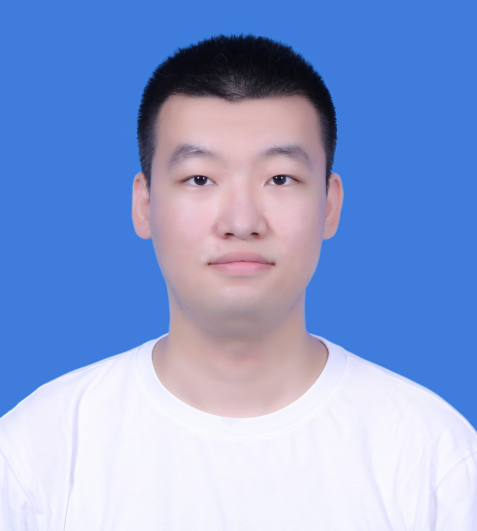}}]
{Jiangxin Sun} received the bachelor's degree in computer science from Sun Yat-sen University in 2020. He is now a M.S. student in the School of Computer Science and Engineering in Sun Yat-sen University. His research interests include instance segmentation and 3D human motion.
\end{IEEEbiography}
\begin{IEEEbiography}[{\includegraphics[width=1in,height=1.25in,clip,keepaspectratio]{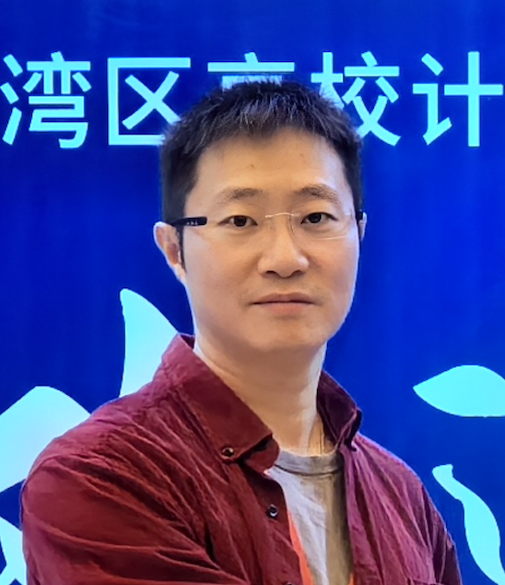}}]
{Dr. Wei-Shi Zheng} is now a full Professor with Sun Yat-sen University. His research interests include person/object association and activity understanding, and the related weakly supervised/unsupervised and continuous learning machine learning algorithms. He has now published more than 200 papers, including more than 150 publications in main journals (TPAMI, IJCV, TIP) and top conferences (ICCV, CVPR, SIGGRAPH, ECCV, NeurIPS). He has ever served as area chairs of ICCV, CVPR, ECCV, BMVC, NeurIPS and etc. He is associate editors/on the editorial board of IEEE-TPAMI, Artificial Intelligence Journal, Pattern Recognition. He has ever joined Microsoft Research Asia Young Faculty Visiting Programme. He is a Cheung Kong Scholar Distinguished Professor, a recipient of the Excellent Young Scientists Fund of the National Natural Science Foundation of China, and a recipient of the Royal Society-Newton Advanced Fellowship of the United Kingdom.
\end{IEEEbiography}
\begin{IEEEbiography}[{\includegraphics[width=1in,height=1.25in,clip,keepaspectratio]{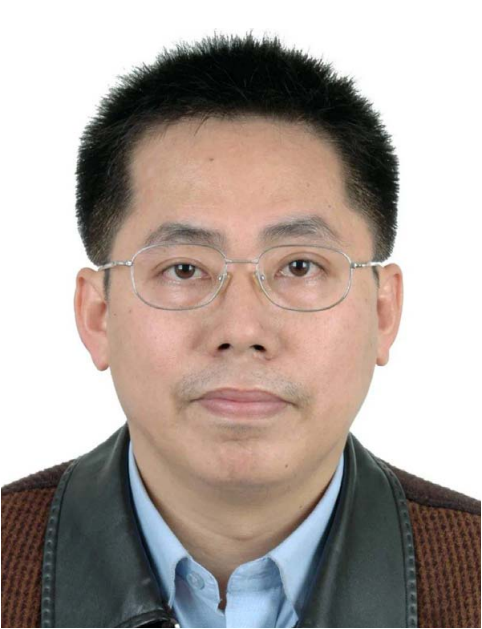}}]
{Jianhuang Lai} (Senior Member, IEEE) received the Ph.D. degree in mathematics from Sun Yat-sen University, China, in 1999. In 1989, he joined Sun Yat-sen University as an Assistant Professor, where he is currently Professor with the School of Computer science and Engineering. His current research interests include the areas of computer vision, pattern recognition, and its applications. He has published over 250 scientific papers in the international journals and conferences on image processing and pattern recognition. He serves as the Deputy Director of the Image and Graphics Association of China.
\end{IEEEbiography}

\vfill

\end{document}